\documentclass[10pt,twocolumn,letterpaper]{article}

\usepackage[pagenumbers]{cvpr} 

\definecolor{cvprblue}{rgb}{0.21,0.49,0.74}
\usepackage[pagebackref,breaklinks,colorlinks,allcolors=cvprblue]{hyperref}
\usepackage{algorithm}
\usepackage{algpseudocode}
\usepackage[dvipsnames,table,xcdraw]{xcolor}
\usepackage{graphicx}
\usepackage{multirow}
\usepackage{dsfont}
\usepackage{subcaption}
\usepackage{multicol}
\usepackage{float}
\usepackage{pifont}

\definecolor{myorange}{rgb}{1, 0.647, 0}
\definecolor{myblue}{rgb}{.118, 0.565, 1}
\usepackage{tocloft} 
\usepackage{titling}
\usepackage{amsthm}
\theoremstyle{remark}

\definecolor{purple}{rgb}{0.65, 0.5, 1.0}
\definecolor{light}{rgb}{0.85, 0.8, 1.0}
\definecolor{white}{rgb} {1.0, 1.0, 1.0}
\definecolor{pink}{rgb}{1.0,0.61372549019,0.70588235294}

\definecolor{orange}{rgb}{1.0000,0.7373,0.5059}
\definecolor{yellow}{rgb}{0.9176, 0.9098, 0.4353}
\definecolor{red}{rgb}{1.0000,0.5686,0.5059}

\title{LIFT: Latent Implicit Functions for Task- and Data-Agnostic Encoding}

\author{
Amirhossein Kazerouni$^{\;\star}$ , \quad Soroush Mehraban$^{\;\star}$, \quad Michael Brudno$^{\;\star}$, \quad Babak Taati$^{\;\star}$ \\[2pt]
{$^{\;\star}$University of Toronto, Vector Institute, University Health Network} \\
{\tt\small \{amirhossein, brudno\}@cs.toronto.edu, soroush.mehraban@mail.utoronto.ca, babak.taati@uhn.ca} \\
\normalsize \url{https://amirhossein-kz.github.io/lift/}
}
\date{}

\begin{document}
\maketitle
\begin{abstract}
Implicit Neural Representations (INRs) are proving to be a powerful paradigm in unifying task modeling across diverse data domains, offering key advantages such as memory efficiency and resolution independence. Conventional deep learning models are typically modality-dependent, often requiring custom architectures and objectives for different types of signals. However, existing INR frameworks frequently rely on global latent vectors or exhibit computational inefficiencies that limit their broader applicability. We introduce \textbf{LIFT}, a novel, high-performance framework that addresses these challenges by capturing multiscale information through meta-learning. LIFT leverages multiple parallel localized implicit functions alongside a hierarchical latent generator to produce unified latent representations that span local, intermediate, and global features. This architecture facilitates smooth transitions across local regions, enhancing expressivity while maintaining inference efficiency. Additionally, we introduce ReLIFT, an enhanced variant of LIFT that incorporates residual connections and expressive frequency encodings. With this straightforward approach, ReLIFT effectively addresses the convergence-capacity gap found in comparable methods, providing an efficient yet powerful solution to improve capacity and speed up convergence. Empirical results show that LIFT achieves state-of-the-art (SOTA) performance in generative modeling and classification tasks, with notable reductions in computational costs. Moreover, in single-task settings, the streamlined ReLIFT architecture proves effective in signal representations and inverse problem tasks. 
\end{abstract}    
\vspace{-1.5em}
\section{Introduction}
\label{sec:intro}
\vspace{-0.5em}
In deep learning, data is traditionally represented by discrete arrays; for example, images are depicted through pixel intensities and 3D shapes via voxel occupancies, both constrained to fixed grid resolutions. However, the underlying signals are inherently continuous, motivating the shift toward representing data with implicit functions. INRs parameterize these continuous functions using neural networks, typically Multilayer Perceptrons (MLPs), to map coordinates directly to signal values. This representation offers advantages in some applications, including handling arbitrary resolutions, efficient memory usage, and versatility across various data modalities such as images~\cite{skorokhodov2021adversarial,sitzmann2020implicit,saragadam2023wire,kazerouni2024incode}, 3D shapes~\cite{park2019deepsdf}, videos~\cite{maiya2024latent,chen2025fast}, and audio~\cite{du2021learning}.

A recent concept, called Functa~\cite{dupont2022data}, describes neural fields as data points modeled using SIRENs— MLPs with sine activations~\cite{sitzmann2020implicit}. Instead of naively representing each neural field with its high-dimensional MLP parameters, Functa adopts a more efficient strategy by sharing parameters across multiple fields and modulating them with per-field biases, a technique known as shift modulation~\cite{dupont2022data}. This approach effectively captures variations among fields with high accuracy while maintaining computational efficiency. In addition to shift modulation, other methods, such as Feature-wise Linear Modulation (FiLM)~\cite{perez2018film} and HyperNetworks~\cite{ha2017hypernetworks}, are employed to manage shared and instance-specific information within neural fields. FiLM-based approaches~\cite{chan2021pi,chan2022efficient} apply simple affine transformations to features, providing simplicity in optimization at the cost of introducing additional parameters. On the other hand, HyperNetworks~\cite{klocek2019hypernetwork,wu2023hyperinr} offers greater flexibility by directly predicting network parameters based on input data. Despite their expressive power, HyperNetworks can suffer from training instability and limited coverage of the data distribution, often requiring dimensionality reduction techniques to mitigate these issues~\cite{maiya2024latent,du2021learning}.

Functa represents data using a single global latent vector, capturing features that span the entire image or data point. Later, mNIF~\cite{you2024generative} constructs generative neural fields by averaging INRs using a shared basis network and weighted modulation. While both methods offer compact representations, they share a key limitation: \textit{the reliance on global latent vectors restricts their ability to capture fine-grained details, reducing their effectiveness in certain downstream tasks like generation and classification.} SpatialFuncta~\cite{bauer2023spatial} employs spatially arranged latent representations, enabling each spatial index to capture localized information. However, this approach remains computationally inefficient. For instance, processing the $32 \times 32$ CIFAR-10 dataset~\cite{krizhevsky2010cifar} with its best-performing classification configuration requires a deep MLP with a depth of 6 and a width of 256, resulting in a high computational cost of 0.271 GFLOPs.

To address these limitations, we propose a novel framework called \textbf{LIFT} (\textbf{L}atent \textbf{I}mplicit \textbf{F}unctions for \textbf{T}ask-agnostic encoding), which captures multiscale information of the data and creates latent modulations that can be applied to a variety of tasks, including generation and classification. LIFT efficiently combines multiple parallel localized implicit functions and a hierarchical latent generator to create a unified latent representation that captures local, intermediate, and global information. Using a meta-learning approach, LIFT efficiently encodes data into a multiscale latent space, enabling smoother transitions between local patches and improving representation expressivity for downstream tasks without sacrificing inference efficiency. Our framework surpasses existing methods by providing a flexible and scalable encoding mechanism that balances local detail and global context. Additionally, LIFT offers a fast encoding approach with a significantly lower FLOP count compared to its counterparts, demonstrating its efficiency in both computational cost and performance. We also introduce \textbf{ReLIFT}, a variant of LIFT, which incorporates residual connections and expressive first-layer frequency scaling to mitigate the convergence-capacity gap often seen in INRs. These enhancements enable the model to capture finer details and represent a broader range of frequencies efficiently. The frequency scaling increases the network’s capacity to learn higher-frequency components, while the residual connections adjust layer weights to amplify higher-order harmonics while preserving fundamental components. This design allows ReLIFT to learn a balance between high and low-frequency information with greater efficiency.  ReLIFT is highly versatile, enabling single-task representations, including images, 3D shapes, audio signals, and inverse problems such as inpainting, denoising, and super-resolution. This highlights the broad applicability of our framework, which enhances representation capabilities compared to existing methods that struggle to capture fine-grained details~\cite{sitzmann2020implicit, tancik2020fourier, fathony2020multiplicative, ramasinghe2022beyond, saragadam2023wire}, manage noise, or generalize across diverse signal types without requiring separate, learnable networks~\cite{mehta2021modulated,kazerouni2024incode}. We summarize the contributions of our work as follows:

\begin{itemize}
    \item We present \textbf{LIFT}, our key contribution- a novel meta-learning framework for task- and data-agnostic latent modulations, capturing multiscale information. LIFT excels across diverse tasks, including generative modeling and classification, without requiring custom architectures.
    \item We introduce a hierarchical latent generator that enables smooth transitions between local patches, facilitating the learning of multilevel information. 
    \item We develop a parallel, localized neural implicit function architecture that balances efficiency and scalability while reducing FLOPs compared to SOTA methods.
    \item We propose ReLIFT, a LIFT variant that integrates residual connections and frequency scaling to accelerate convergence speed. Tailored for single-task applications such as signal representation and inverse problems, it reduces the convergence-capacity gap compared to SOTA models.

\end{itemize}

\vspace{-0.75em}
\section{Related Work}
\label{sec:formatting}

\noindent \textbf{Implicit Neural Representations} model continuous signals by mapping coordinates to signal values using neural networks. Recent advances in INRs have addressed the challenge of spectral bias in ReLU-based MLPs, which struggle to capture high-frequency details~\cite{rahaman2019spectral}. SIREN~\cite{sitzmann2020implicit} introduced sinusoidal activations to capture high-frequency components effectively, while Tancik et al.~\cite{tancik2020fourier} employed Fourier feature mappings to project input coordinates into a higher-dimensional space, thereby enhancing performance. MFN~\cite{fathony2020multiplicative} further improved signal representation by incorporating specialized filters. WIRE~\cite{saragadam2023wire} introduced a Gabor wavelet-based nonlinearity to reduce artifacts and enhance representation accuracy. Despite these advancements, INRs still face challenges in scaling to large datasets, as they are primarily optimized for single-data representations. Even for single-tasks, their potential is often limited to fully exploit the high representation capacity of INRs, resulting in suboptimal learning of fine-grained details. 
\begin{figure*}[t]
    \centering
    \includegraphics[width=1\textwidth]{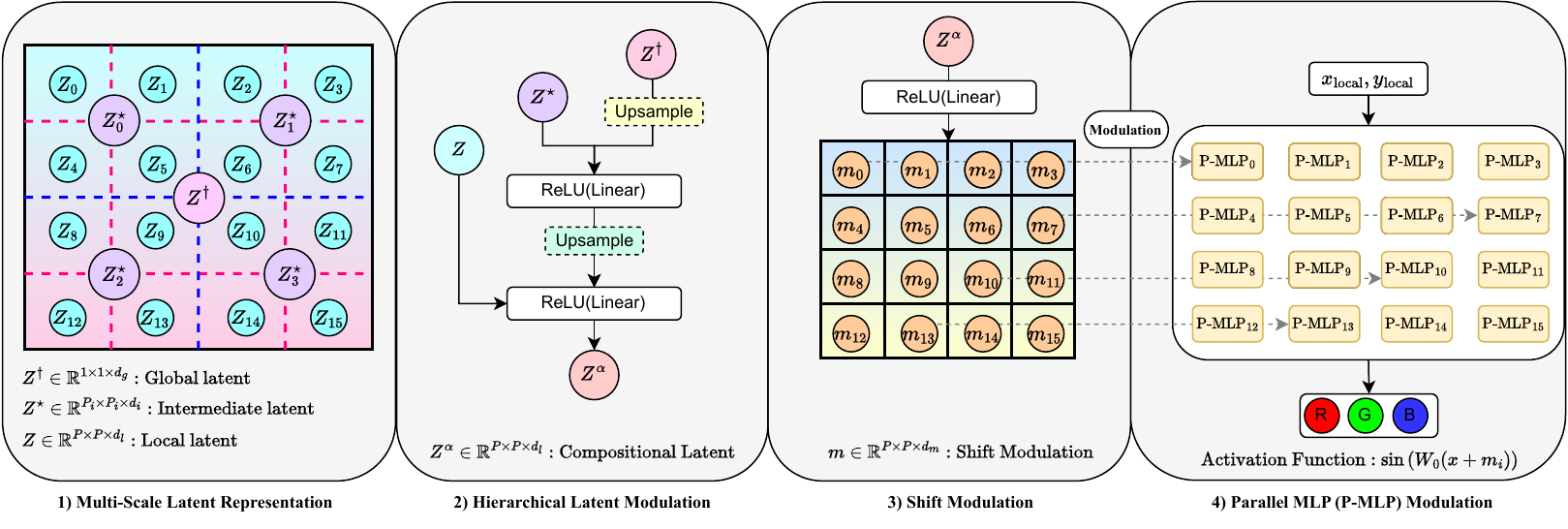}
    \vspace{-1em}
    \caption{Overview of the \textbf{LIFT} architecture, illustrating its progressive latent modulation stages. (1) A hierarchy of learnable latents—global, intermediate, and local—is initialized to capture multiscale features. (2) These latents are integrated through hierarchical latent modulation, ensuring smooth transitions across local patch boundaries by sharing information between regions. (3)  Shift modulations are produced for each Parallel MLP (P-MLP). (4) P-MLPs, modulated by their respective shift parameters and utilizing sinusoidal activation functions, generate high-fidelity reconstructions.}
    \label{fig:main}
    \vspace{-1.3em}
\end{figure*}

\noindent \textbf{Generalized Implicit Neural Representations} extend traditional INRs to handle more complex tasks and larger datasets. Functa~\cite{dupont2022data} introduces global data-specific bias vectors for efficient neural network modulation. SpatialFuncta~\cite{bauer2023spatial} refines the Functa framework by spatially structuring latent neural field representations to improve scalability. However, it relies on a large MLP, resulting in high computational costs and making it challenging to extend to 3D domains. Also, relying solely on local information can hinder tasks such as classification, where global context is crucial. HyperDiffusion~\cite{erkocc2023hyperdiffusion} eliminates the need for latent vectors by directly sampling neural field weights via diffusion, reducing computational costs while preserving quality. However, it is limited to voxel-based domains and struggles to generalize to more complex modalities like images. GEM~\cite{du2021learning} generates instance-specific neural networks using hypernetworks but incurs high computational overhead due to its reliance on manifold learning constraints, which require constant access to training data. GASP~\cite{dupont2022generative} also employs hypernetworks but suffers from training instability. DPF~\cite{zhuang2023diffusion} introduces a diffusion-based approach for explicit field parametrization, achieving high fidelity at the cost of increased computational resources. mNIF~\cite{you2024generative} proposes a flexible approach by averaging basis networks with learned modulation. Our approach extends previous methods by introducing efficient hierarchical modulations that integrate multiscale information, enhancing performance in both generative and classification tasks while remaining computationally efficient. Our goal is to address challenges in both large-scale datasets and single-data scenarios through our proposed methods, LIFT and ReLIFT.
\vspace{-0.5em}
\section{Method}
\vspace{-0.5em}
In this section, we introduce LIFT, a novel two-stage framework designed to efficiently encode input data into a multiscale latent space and use these latents in tasks like generation and classification (\autoref{fig:main}). LIFT combines parallel localized implicit functions and a hierarchical latent generator to create a unified latent representation that captures local, intermediate, and global information. The entire process is trained in a meta-learning approach inspired by~\cite{dupont2022data}. Here, we outline the key components of LIFT, from the modulation of implicit functions to the design of the latent generator and the optimization process.

\vspace{-0.25em}
\subsection{Localized Neural Implicit Functions}
\vspace{-0.5em}
INRs model continuous signals across various modalities by learning a function \( \mathcal{F}_{\theta}: \mathbb{R}^D \rightarrow \mathbb{R}^C \) that maps input coordinates \( \mathbf{x} = (x_1, x_2, \dots, x_D) \in \mathbb{R}^D \) to signal values \( S(\mathbf{x}) \in \mathbb{R}^C \) using a neural network. For a given signal \( S \), defined over a discrete grid of size \( N_1 \times N_2 \times ... \times N_{D} \) along each dimension, our goal is to find \( \mathcal{F}_{\theta} \) such that:
\begin{equation}
\small
\left\| \mathcal{F}_{\theta}(x_1^{i_1}, x_2^{i_2}, \dots, x_D^{i_D}) - S[i_1, i_2, \dots, i_D] \right\| = 0,
\end{equation}
for all \( i_d = 1, 2, \dots, N_d \) and \( d = 1, 2, \dots, D \), where the normalized coordinates \( x_d^{i_d} \) are computed as:
$x_d^{i_d} = \frac{i_d - 1}{N_d - 1}$, scaling the index \( i_d \) to the interval \([0, 1]\). The dataset \( \mathcal{D} \) consists of coordinate-value pairs:
\footnotesize
\begin{equation}
    \mathcal{D} = \left\{ \left( \mathbf{x}^{\mathbf{i}}, \, S[\mathbf{i}] \right) \ \Big| \ \mathbf{i} = (i_1, i_2, \dots, i_D), \ i_d = 1, 2, \dots, N_d \right\}.
\end{equation}
\normalsize
Existing methods (e.g.,~Functa~\cite{dupont2022data,bauer2023spatial}) approximate \( \mathcal{F}_{\theta} \) using an MLP trained on \( \mathcal{D} \), mapping the input coordinates directly to signal values. To enhance computational efficiency and scalability, we express \( \mathcal{F}_{\theta} \) as a sum of sub-functions:
\begin{equation}
    \small
    \mathcal{F}_{\theta}(\mathbf{x}) = \sum_{m=1}^{M^D} f_m(\mathbf{x}) \cdot \mathds{1}_m(\mathbf{x}),
\end{equation}
where \( f_m: \mathbb{R}^D \rightarrow \mathbb{R}^C \) are continuous functions modeling the signal within localized regions, and \( \mathds{1}_m: \mathbb{R}^D \rightarrow \{0, 1\} \) are indicator functions defined as:
\begin{equation}
    \small
    \mathds{1}_{m}(\mathbf{x}) =
    \begin{cases}
    1, & \text{if } \mathbf{x} \in D_m, \\
    0, & \text{otherwise},
    \end{cases}
\end{equation}
with \( \{ D_m \}_{m=1}^{M^D} \) forming a partition of the domain \([0, 1]^D\). Each region \( D_m \) is defined by the bounds:
\begin{equation}
    \small
    D_m = \left\{ \mathbf{x} \in \mathbb{R}^D \ \bigg| \ a_d^{(k_d)} < x_d \leq b_d^{(k_d)}, \ \forall d = 1, 2, \dots, D \right\},
\end{equation}
where $a_d^{(k_d)} = \frac{k_d - 1}{M}, \quad
b_d^{(k_d)} = \frac{k_d}{M}$, and \( k_d = 1, 2, \dots, M \) is the region index along dimension \( d \). The total number of regions is \( M^D \), and the mapping from \( m \) to multi-dimensional indices \( (k_1, k_2, \dots, k_D) \) is given by:

\begin{equation}
    \small
    k_d = \left\lfloor \frac{(m - 1)}{M^{D - d}} \right\rfloor \bmod M + 1.
\end{equation}
By partitioning the domain into regions \( D_m \), each sub-function \( f_m \) focuses on modeling the signal within its corresponding region, allowing for localized learning and improved computational efficiency. Consider the specific case of modeling a 3D shape, where \( D = 3 \) and \( C = 1 \). The signal \( S \)
 represents a scalar field over a 3D grid, such as an occupancy function or a signed distance function (SDF).  It maps 3D coordinates  \( \mathbf{x} = (x, y, z) \)  to scalar values \( S(\mathbf{x}) \). Our goal is to find \( \mathcal{F}_{\theta}: \mathbb{R}^3 \rightarrow \mathbb{R} \) such that:
\begin{equation}
\small
    \left\| \mathcal{F}_{\theta}(x^{i}, y^{j}, z^{k}) - S[i, j, k] \right\| = 0,
    \tag{10}
\end{equation}
for all \( i, j, k = 1, 2, \dots, N \), with normalized coordinates:
\begin{equation}
\small
    x^{i} = \frac{i - 1}{N - 1}, \quad y^{j} = \frac{j - 1}{N - 1}, \quad z^{k} = \frac{k - 1}{N - 1}.
    \tag{11}
\end{equation}
The dataset \( \mathcal{D} \) consists of coordinate-value pairs:
\begin{equation}
\small
    \mathcal{D} = \left\{ \left( x^{i}, y^{j}, z^{k}, \, S[i, j, k] \right) \mid i, j, k = 1, 2, \dots, N \right\}.
    \tag{12}
\end{equation}
We express \( \mathcal{F}_{\theta}(x, y, z) = \sum_{m=1}^{M^3} f_m(x, y, z) \cdot \mathds{1}_m(x, y, z) \), where each sub-function \( f_m \) models the signal within its corresponding region \( D_m \) defined in 3D space. 

\vspace{-0.25em}
\subsection{Hierarchical Latent Generator (HLG)}
\vspace{-0.5em}
To efficiently capture localized variations in the signal \( S(\mathbf{x}) \) and to mitigate discontinuities at region boundaries, we propose a \textbf{Hierarchical Latent Generator}. Specifically, we partition the domain \([0,1]^D\) into \( M^D \) regions \( \{ D_m \}_{m=1}^{M^D} \) as previously defined. For each region \( D_m \), we assign an independent small MLP \( f_m \), collectively forming the set \( \{ f_m \}_{m=1}^{M^D} \). Each local MLP is then modulated by its corresponding latent vector. 

\noindent\textbf{Modulation of Local MLPs:} The modulation is applied via a modulation function $\mathcal{M}$ (a linear layer) on a compositional latent $Z^{\alpha}$ (see~\autoref{fig:main}), which generates modulation parameters for MLP layers, $m = \mathcal{M}(Z^{\alpha})$. Specifically, the modulation parameters \( \mathbf{m}_m^{(l)} \) are generated to modulate the activation functions of the MLP layers, allowing each MLP to adapt its representation based on regional characteristics. The activations for each MLP \( f_m \) are computed as:
\begin{equation}
\small
    \mathbf{h}_m^{(l)} = \sin\left( \omega_0 \left( \mathbf{W}_m^{(l)} \mathbf{h}_m^{(l-1)} + \mathbf{b}_m^{(l)} + \mathbf{m}_m^{(l)} \right) \right), 
\end{equation}
where \( l = 1, 2, \dots, L-1 \), \( \mathbf{h}_m^{(0)} = \mathbf{x}_m \in [0,1]^D \) is the local input coordinate vector, \( \omega_0 \in \mathbb{R}^{+} \) is a fixed positive scalar regulating the frequency of the model, and \( \mathbf{W}_m^{(l)} \in \mathbb{R}^{W \times W} \), \( \mathbf{b}_m^{(l)} \in \mathbb{R}^{W} \) are the weights and biases of \( f_m \) for layer \( l \).  The output of each MLP \( f_m(\mathbf{x}) \) is defined as: $f_m(\mathbf{x}) = \mathbf{W}_m^{(L)} \mathbf{h}_m^{(L-1)} + \mathbf{b}_m^{(L)},$ where \( \mathbf{W}_m^{(L)} \in \mathbb{R}^{C \times W} \) and \( \mathbf{b}_m^{(L)} \in \mathbb{R}^C \) are the final layer weights and biases specific to \( D_m \). We also introduce a variant of our SIREN-based modulation activation function, termed \texttt{ReLIFT}, which incorporates an intuitive residual connection after the activation function and applies a frequency scaling factor.

\noindent\textbf{Remark.} Let a standard SIREN~\cite{sitzmann2020implicit} computes the first-layer output as $\mathbf{z}^{(0)} = \sin\left(\Omega \mathbf{r}\right)$, where \(\Omega = \omega_0 W^{(0)} \in \mathbb{R}^{T}\). Considering a three-layer SIREN, the network can be defined as:
\small
\begin{equation}
    f(\mathbf{r}; \theta) = \mathbf{w}^{(2)\top} \sin\left( \mathbf{W}^{(1)} \sin\left( \Omega \mathbf{r} \right) \right),
\end{equation}
\normalsize
with \(\mathbf{W}^{(1)} \in \mathbb{R}^{F \times T}\) and \(\mathbf{w}^{(2)} \in \mathbb{R}^F\). The input to each neuron in the second layer is a linear combination of sinusoids at frequencies determined by \(\Omega\). The output of a neuron \(z_m^{(1)}\) in the second layer can be written as:
\small
\begin{align}
    z_m^{(1)} &= \sin\left( \sum_{t=0}^{T-1} W_{m,t}^{(1)} \sin\left( \boldsymbol{\omega}_t^\top \mathbf{r} \right) \right),
\end{align}
\normalsize
where \(\boldsymbol{\omega}_t^\top\) denotes the \(t\)-th row of \(\Omega\). Inspired by ~\cite{yuce2022structured}, for a three-layer SIREN, we can express the output using the Fourier-Bessel expansion:
\small
\begin{equation}
    \begin{aligned}
    f(\mathbf{r}; \theta) = &\sum_{m=0}^{F-1} w_m^{(2)} z_m^{(1)} = \sum_{m=0}^{F-1} w_m^{(2)} \\
    &\sum_{\mathbf{s} \in \mathbb{Z}^T} \left( \prod_{t=0}^{T-1} J_{s_t}\left( W_{m,t}^{(1)} \right) \right) 
    \sin\left( \sum_{t=0}^{T-1} s_t \boldsymbol{\omega}_t^\top \mathbf{r} \right).
    \end{aligned}
\end{equation}
\normalsize
This expression indicates that the network output is a sum of sinusoidal functions at frequencies \( \sum_{t=0}^{T-1} s_t \boldsymbol{\omega}_t \), where \( s_t \in \mathbb{Z} \). This implies that by scaling \( \sum_{t=0}^{T-1} s_t \boldsymbol{\omega}_t \), we can increase the network’s capacity to learn higher-frequency components. In addition, the coefficients of these sinusoids are determined by the products of Bessel functions \( J_{s_t}\left( W_{m,t}^{(1)} \right) \) and the weights \( w_m^{(2)} \). Due to the properties of Bessel functions, which generally decrease in magnitude with increasing order \( |s_t| \) when the argument \( W_{m,t}^{(1)} \) is small, higher-order harmonics (those with larger \( |s_t| \)) tend to have smaller coefficients. This results in an implicit bias towards lower-frequency components, concentrating most of the output signal's energy around the fundamental frequencies \( \boldsymbol{\omega}_t \). However, increasing the scale of inner layer coefficients, like \( W^{(1)} \), boosts higher-order harmonics, allowing the network to capture a wider frequency range. Therefore, we introduce \texttt{ReLIFT}, which \textbf{(1)} scales the input layer frequencies to represent higher-frequency components and \textbf{(2)} adds a residual connection to refine \( W^{(1)} \), amplifying harmonics while preserving fundamentals.  Consider a three-layer network where the first layer is defined as
\small
\begin{equation}
    \mathbf{z}^{(0)} = \sin\bigl(\textcolor{blue}{\underline{\gamma}}\,\Omega\,\mathbf{r}\bigr).
\end{equation}
\normalsize
Then, the second layer becomes
\small
\begin{equation}
    \mathbf{z}^{(1)} = \sin\Bigl(\mathbf{W}^{(1)}\sin\bigl(\gamma\,\Omega\,\mathbf{r}\bigr)\Bigr) \textcolor{blue}{\underline{+ \sin(\gamma\,\Omega\,\mathbf{r}\bigr)}}.
\label{eq:layer2}
\end{equation}
\normalsize
By applying the Fourier-Bessel expansion, we can show that
\small
\begin{align}
    \label{eq:second_layer_final}
    z_m^{(1)} &= \underbrace{\sum_{\mathbf{s} \in \mathbb{Z}^T} \left( \prod_{t=0}^{T-1} J_{s_t}\left( W_{m,t}^{(1)} \right) \right) \sin\!\left( \gamma \sum_{t=0}^{T-1} s_t \, \boldsymbol{\omega}_t^\top \mathbf{r} \right)}_{\text{Higher-order harmonics}} \nonumber \\
     &\quad + \underbrace{\sin \left( \gamma \omega_{m}^{T} r\right)}_{\text{Fundamental component}}.
\end{align}
\normalsize
When \( \gamma > 1 \) increases, the frequencies \( \gamma \sum_{t=0}^{T-1} s_t \boldsymbol{\omega}_t \) increase proportionally, extending the network's capacity to represent higher-frequency components by broadening the range of frequencies it can model. The residual connection ensures the robust representation of lower-order harmonics (fundamental frequencies). Although the Bessel expansion contributes to higher-order harmonics, the residual connection allows the network to avoid depending solely on the Bessel functions to capture lower frequencies. This balance allows the network to adjust \( W^{(1)} \) to amplify higher-order harmonics while preserving fundamental components. See the Supplementary Material (SM) for proof and analysis.


\noindent\textbf{Hierarchical Latent Modulation:} Prior works~\cite{chan2021pi,dupont2022data} apply global modulations to INRs by mapping a low-dimensional latent vector to modulation parameters for the entire model. Building on~\cite{dupont2022data}, where modulation is applied as a bias vector, we introduce multiscale modulations to capture the hierarchical information of the input signal. Without loss of generality, assuming a square image of size \( N \times N \) partitioned into \( P \times P \) patches, we assign each patch an independent local P-MLP for reconstruction. This results in \( P^2 \) patches and a latent space of shape \( P \times P \times {d_{l}} \). However, modulating each P-MLP solely based on its local latent vector can cause discontinuities at patch boundaries due to a lack of shared information between neighboring regions. Therefore, we propose a multiscale latent representation by defining three levels of latents: \textbf{1)~Global Latent} \( \mathbf{Z}^{\dagger} \in \mathbb{R}^{1 \times 1 \times d_g} \)(captures global features of the signal), \textbf{2)~Intermediate Latent} \( \mathbf{Z}^{\star} \in \mathbb{R}^{{P_i} \times {P_i} \times d_i} \)(encodes intermediate-scale feature), and \textbf{3)~Local Latent} \( \mathbf{Z} \in \mathbb{R}^{{P} \times {P} \times d_l} \)(provides fine-grained local details.) We formalize the hierarchical latent modulation as a series of transformations: First, global to intermediate latent transformation:
\begin{equation}
\small
    \mathbf{Z'} = \text{Linear}_1\left( \text{Concat}\left( \text{Upsample}\left( \mathbf{Z}^{\dagger}, {P_i}, {P_i} \right), \mathbf{Z}^{\star} \right) \right),
\end{equation}
where \( \text{Upsample}(\cdot) \) increases the spatial resolution to \( {P_i} \times {P_i} \), and \( \text{Linear}_1(\cdot) \) applies a linear transformation mapping to the intermediate latent space. Then, intermediate to local latent transformation:
\begin{equation}
\small
    \mathbf{Z}^{\alpha} = \text{Linear}_2\left( \text{Concat}\left( \text{Upsample}\left( \mathbf{Z'} , {P}, {P} \right), \mathbf{Z} \right) \right).
\end{equation}
The compositional latent \( \mathbf{Z}^{\alpha} \) encodes information from multiple scales and is used for modulation. This hierarchical structure allows global latent information to influence local latents, ensuring that neighboring regions share high-level representations. As a result, transitions between patches are smoother, reducing boundary discontinuities. For example, the local latents \( \mathbf{Z}_m^{\alpha} \) and \( \mathbf{Z}_n^{\alpha} \) of adjacent regions \( D_m \) and \( D_n \) are derived from the same intermediate latent \( \mathbf{Z}^{\star} \), which itself originates from the global latent \( \mathbf{Z}^{\dagger} \). This shared hierarchical information ensures consistent representations across regions and smooth transitions.

\subsection{LIFT Optimization}
The training of our proposed LIFT framework is executed in two sequential stages: \textbf{Context Adaptation} and \textbf{Task-Driven Generalization}. This hierarchical approach ensures that the latent modulations \( Z^{\alpha} \) gather multiscale information, facilitating their use in downstream tasks such as classification and generation while maintaining a compact and efficient neural field representation.
\vspace{-1em}
\subsubsection{Stage 1: Context Adaptation}
\vspace{-0.5em}
In this phase, we employ a meta-learning strategy to generate a comprehensive dataset of latent modulations. Specifically, we optimize multiscale latents to store the latent modulation \( Z^{\alpha} \), which encapsulates multiscale information and enables more efficient downstream task performance compared to using global, intermediate, and local latent modulations separately or collectively. This stage leverages a meta-learning framework inspired by fast context adaptation via meta-learning (CAVIA) \cite{zintgraf2019fast}, wherein:

\noindent\ding{202} \textbf{Inner Loop}: Focuses on updating the multiscale latent modulations \( Z^{\dagger}, Z^{\star},\) and $Z$ using the stochastic gradient descent (SGD) optimizer for $T_{\text{inner}}$ steps.

\noindent\ding{203} \textbf{Outer Loop}: Updates network weights, keeping them fixed during inner-loop using Adam optimizer~\cite{kingma2014adam}.

\noindent We optimize the model with a  loss function that includes:

\noindent \textbf{1) Reconstruction Loss}: Measures the mean squared error (MSE) between the reconstructed output and input signal:
\begin{equation*}
\small
    \mathcal{L}_{\text{Rec}} = \frac{1}{|\mathcal{D}|} \sum_{\mathbf{x} \in \mathcal{D}} \left\| \mathcal{F}_{\theta}(\mathbf{x}) - S(\mathbf{x}) \right\|_2^2,
\end{equation*}
where \( \mathcal{D} \) denotes the dataset of coordinate-value pairs.

\noindent \textbf{2) Latent Smoothness Loss}: Encourages a structured latent space by promoting local consistency among the learned latent vectors. For each latent \( Z^{\alpha}_m \), loss is computed as:
\begin{equation*}
\small
    \mathcal{L}_{\text{Smoothness}}(Z^{\alpha}_m) = \frac{1}{K} \sum_{k=1}^K \| Z^{\alpha}_m - Z_{k}^{\alpha} \|_2^2,
\end{equation*}
where \( Z_{k}^{\alpha} \) represents the \(k \)-th nearest neighbor of \( Z^{\alpha}_m \), and \( K \) is a hyperparameter specifying the number of neighbors. The total loss function is therefore $\mathcal{L}_{\text{Total}} = \mathcal{L}_{\text{Rec}} + \lambda \mathcal{L}_{\text{Smoothness}}$, where \( \lambda \) balances the contribution of the reconstruction and smoothness losses. This dual-objective optimization ensures that the latent modulations can reconstruct the input signal and maintain a smooth latent space, which is crucial for downstream task performance.

\begin{table*}[t]
    \centering
    \vspace{-0.5em}
    \caption{Image and voxel reconstruction and generation performance on CelebA-HQ \(64^2\) and ShapeNet \(64^3\). The results are from the corresponding papers and mNIF~\cite{you2024generative}. Each cell is color-coded to represent the \colorbox{purple}{\kern-\fboxsep best\kern-\fboxsep} and \colorbox{light}{\kern-\fboxsep second-best\kern-\fboxsep} performance. \textit{Learn.} and \textit{Inf.} indicate learnable and inference, respectively.}
    \vspace{-0.75em}
    \resizebox{\textwidth}{!}{%
    \begin{tabular}{l|cc|cccc|cc|c||cc|cc|cc|c}
    \toprule
    \multirow{3}{*}{Model} 
        & \multicolumn{9}{c||}{CelebA-HQ \(64^2\)} 
        & \multicolumn{7}{c}{ShapeNet \(64^3\)} \\ 
    \cmidrule(lr){2-10} \cmidrule(lr){11-17}
        & \multicolumn{2}{c|}{Reconstruction} 
        & \multicolumn{4}{c|}{Generation} 
        & \multicolumn{2}{c|}{\# Params} 
        & \multirow{2}{*}{\# FLOPs $\downarrow$} 
        & \multicolumn{2}{c|}{Reconstruction} 
        & \multicolumn{2}{c|}{Generation} 
        & \multicolumn{2}{c|}{\# Params} 
        & \multirow{2}{*}{\# FLOPs $\downarrow$} \\
    \cmidrule(lr){2-3} \cmidrule(lr){4-7} \cmidrule(lr){8-9} \cmidrule(lr){11-12} \cmidrule(lr){13-14} \cmidrule(lr){15-16}
        & PSNR $\uparrow$ 
        & rFID $\downarrow$ 
        & FID $\downarrow$ 
        & Precision $\uparrow$ 
        & Recall $\uparrow$ 
        & F1 $\uparrow$ 
        & Learn. (M) 
        & Inf. (K) 
        & 
        & MSE $\downarrow$ 
        & PSNR $\uparrow$ 
        & Coverage $\uparrow$ 
        & MMD $\downarrow$ 
        & Train. (M) 
        & Inf. (K) 
        & \\
    \midrule
    Functa \cite{dupont2022data} 
        & 26.6 & 28.4 
        & 40.4 & 0.577 & 0.397 & 0.470 
        & 3.3 & 2629.6 & 8.602 G 
        & -- & -- 
        & -- & -- 
        & -- & -- & -- \\
    GEM \cite{du2021learning} 
        & --   & -- 
        & 30.4 & 0.642 & \cellcolor{white}0.502 & \cellcolor{white}0.563 
        & 99.0 & 921.3 & 3.299 G 
        & \cellcolor{light}0.0153 & \cellcolor{white}21.3 
        & 0.409 & \cellcolor{light}0.0014 
        & 99.0 & 921.3 & 207.0 G \\
    GASP \cite{dupont2022generative} 
        & --   & -- 
        & \cellcolor{white}13.5 & 0.836 & 0.312 & 0.454 
        & 34.2 & 83.1 & 305 M 
        & 0.0296 & 16.5 
        & 0.341 & 0.0021 
        & 34.2 & 83.1 & 8.7 G \\
    DPF \cite{zhuang2023diffusion} 
        & --   & -- 
        & \cellcolor{light}13.2 & \cellcolor{white}0.866 & 0.347 & 0.495 
        & 62.4 & -- & -- 
        & -- & -- 
        & \cellcolor{white}0.419 & \cellcolor{white}0.0016 
        & 62.4 & -- & -- \\
    mNIF-S \cite{you2024generative} 
        & \cellcolor{white}31.5 & \cellcolor{white}10.9 
        & 21.0 & 0.787 & 0.324 & 0.459 
        & 4.6 & 17.2 & 69 M 
        & \cellcolor{white}0.0161 & 20.9 
        & \cellcolor{light}0.430 & \cellcolor{light}0.0014 
        & 4.6 & 17.2 & 4.4 G \\
    mNIF-L \cite{you2024generative} 
        & \cellcolor{light}34.5 & \cellcolor{light}5.8 
        & \cellcolor{light}13.2 & 0.902\cellcolor{purple} & \cellcolor{light}0.544 & \cellcolor{light}0.679 
        & 33.4 & 83.3 & 340 M 
        & 0.0166 & \cellcolor{light}21.4 
        & \cellcolor{purple}0.437 & \cellcolor{purple}0.0013 
        & 46.3 & 83.3 & 21.6 G \\
        \midrule
    \textbf{LIFT} 
        & 39.4\cellcolor{purple} & 2.6\cellcolor{purple} 
        & 10.0\cellcolor{purple} & \cellcolor{light}0.900 & 0.631\cellcolor{purple} & \cellcolor{purple}0.742 
        & 0.9 & 839.0 & 54.52 M 
        & \cellcolor{purple}0.00053 & \cellcolor{purple}35.2 
        & \cellcolor{light}0.430 & \cellcolor{white}0.0016 
        & 8.8 & 8185.0 & 4.38 G \\
    \bottomrule
    \end{tabular}
    }
\label{tab:2d_img}
\vspace{-1em}
\end{table*}
\vspace{-1em}
\subsubsection{Stage 2: Task-Driven Generalization}
\vspace{-0.5em}
Post-meta-learning, we construct a modulation dataset by iteratively applying the inner loop to each dataset item, using $T_{\text{inner}}$ gradient steps for effective modulation fitting. This process is replicated across both the training and test datasets. For generative tasks, we use DDPM~\cite{ho2020denoising} coupled with a custom UNet during training and DDIM~\cite{song2021denoising} for sample generation during inference. For classification, we train VMamba~\cite{zhu2024vision} on the LIFT representation dataset.

\vspace{-0.75em}
\vspace{-0.5em}
\section{Experiments}
We conduct our experiments on both image and voxel datasets to demonstrate the versatility and scalability of our approach. For image tasks, we use CIFAR-10 \(32^2\)~\cite{krizhevsky2010cifar}, CelebA-HQ \(64^2\)~\cite{karras2017progressive}, and ImageNet-100 \(256^2\)~\cite{russakovsky2015imagenet} datasets, evaluating reconstruction quality using metrics such as PSNR and rFID, and evaluating generation quality with FID~\cite{heusel2017gans}, precision, recall, and F1 scores \cite{sajjadi2018assessing,
naeem2020reliable}. For voxel-based tasks, we employ the ShapeNet \(64^3\) dataset~\cite{chang2015shapenet}, measuring reconstruction performance with MSE and PSNR, and generation quality using coverage and MMD. Our models are implemented in JAX~\citep{bradbury2018jax} with Haiku~\citep{haiku2020github} and trained using a meta-learning framework on A40 GPUs. The training details for both stages—Context Adaptation and Task-Driven Generalization—including detailed descriptions of the datasets, model configurations, and training protocols, are provided in SM.

\vspace{-0.25em}
\subsection{2D Analysis}
\begin{figure*}[t]
    \centering
    \begin{minipage}[t]{0.34\linewidth}
        \centering
        \includegraphics[width=1\linewidth]{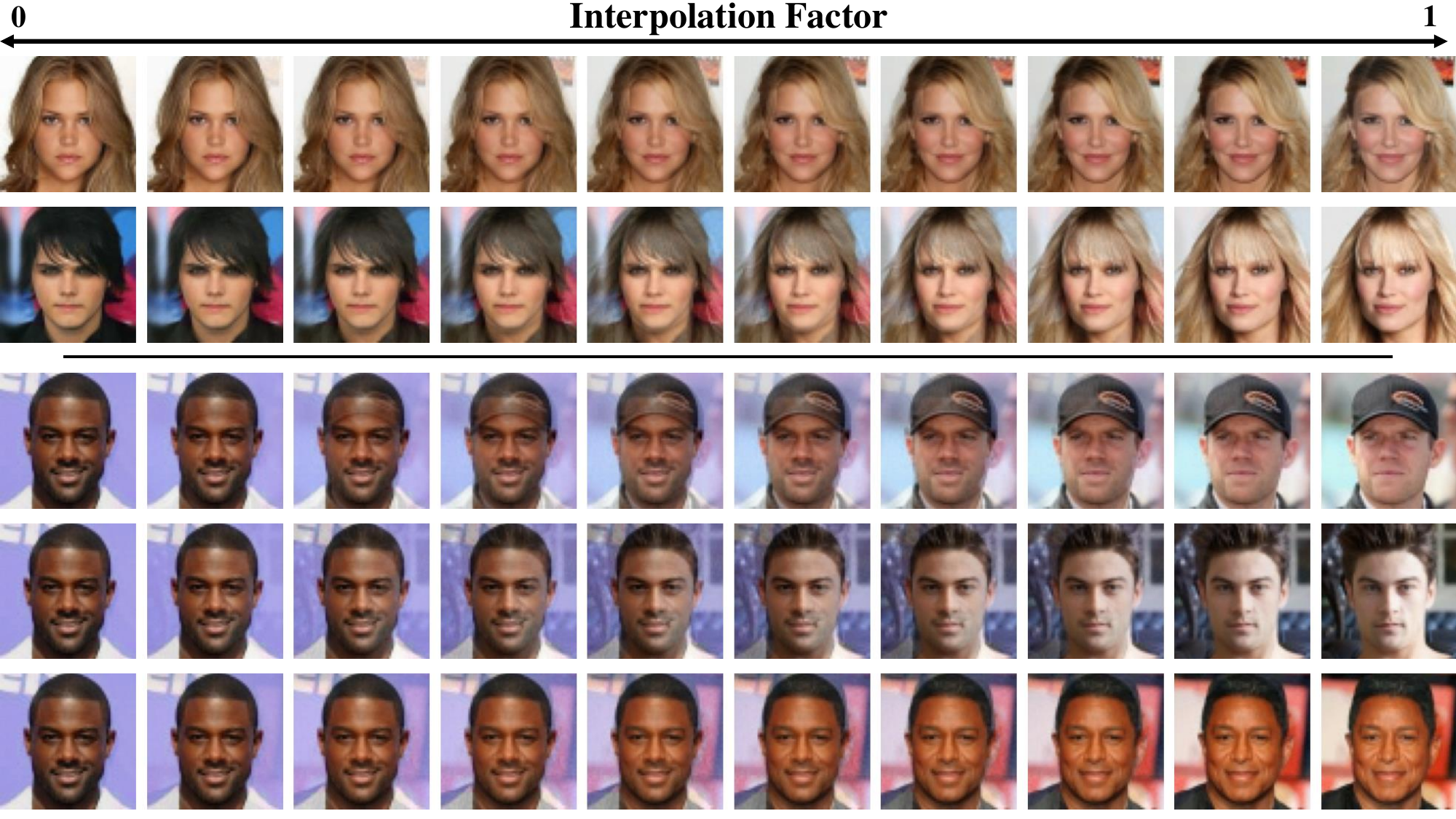}
        \caption{2D latent interpolation showing gradual attribute changes (zoom in for details).}
        \label{fig:2d_interpolation}
    \end{minipage}
    \hfill
    \begin{minipage}[t]{0.65\linewidth}
        \centering
        \includegraphics[width=1\linewidth]{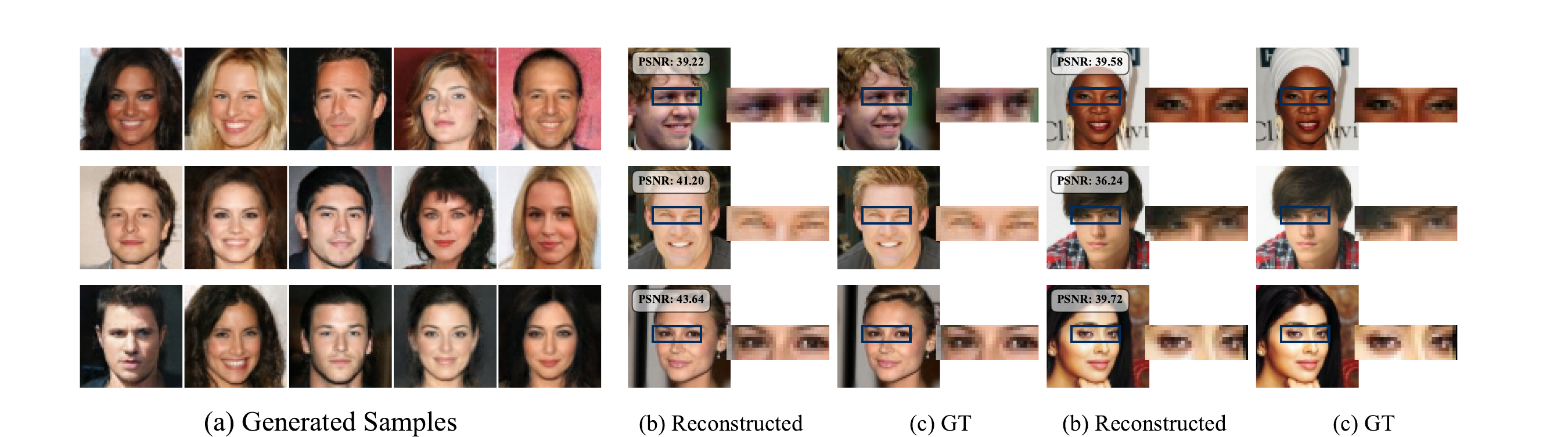}
        \caption{Generated samples along with reconstructed and corresponding ground truth images, with a zoomed-in view for detail.}
        \label{fig:2d_image}
    \end{minipage}
    \begin{minipage}[t]{0.34\linewidth}
        \centering
        \includegraphics[width=1\linewidth]{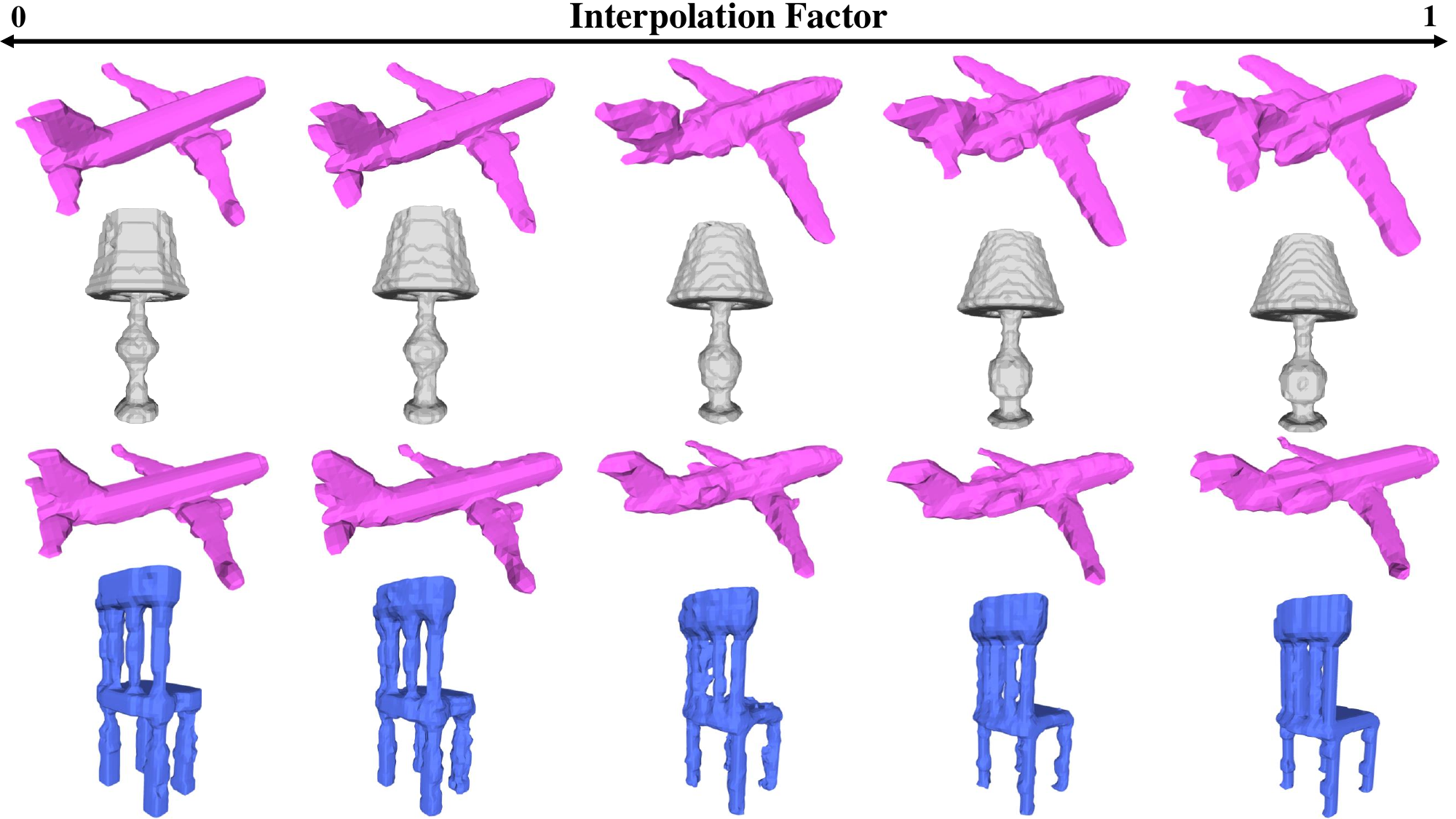}
        \caption{3D latent interpolation showing continuous object shape transformations.}
        \label{fig:3d_interpolation}
    \end{minipage}
    \hfill
    \begin{minipage}[t]{0.65\linewidth}
        \centering
        \includegraphics[width=1\linewidth]{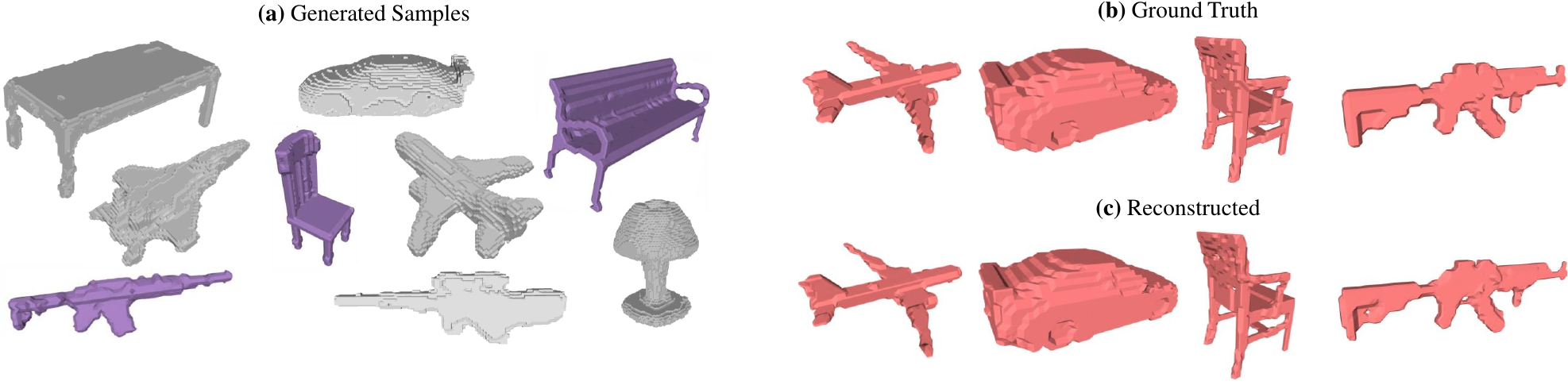}
        \caption{Generated, reconstructed, and ground truth 3D shapes.}
        \label{fig:3d_image}
    \end{minipage}
    \vspace{-1.5em}
\end{figure*}

\vspace{-0.25em}
\noindent\textbf{Reconstruction:}
\autoref{tab:2d_img} presents the reconstruction performance on the CelebA-HQ dataset. LIFT achieves a PSNR of 39.4\,dB, significantly surpassing the previous best result of 34.5\,dB by mNIF-L. This improvement of over 4.9 indicates a substantial enhancement in reconstruction fidelity. Additionally, LIFT attains an rFID of 2.6, markedly lower than mNIF-L's 5.8, demonstrating closer alignment with the real data distribution. The superior reconstruction quality can be attributed to LIFT's efficient hierarchical latent representations, which capture both global structure and fine-grained details. Unlike models like Functa, which have higher parameter counts but lower performance, LIFT demonstrates that a well-designed latent space can lead to better reconstructions with significantly fewer parameters. In the ImageNet-100 dataset with higher resolution, LIFT shows robust performance in just 250K training iterations (see~\autoref{tab:imagenet}). This consistency across resolutions highlights LIFT's scalability and robustness in handling complex, high-resolution data. \autoref{fig:2d_image} presents qualitative reconstructions where LIFT's outputs exhibit sharp details and high visual fidelity, closely resembling the originals, which supports the quantitative metrics and emphasizes its effectiveness in representing high-fidelity information.
\begin{table}[t]
    \centering
    \vspace{-0.5em}
    \caption{Image reconstruction performance on ImageNet-100 \(256^2\) trained for 250K iterations.}
    \vspace{-0.5em}
    \resizebox{0.85\linewidth}{!}{%
    \begin{tabular}{ccccccc}
    \toprule
    \multicolumn{3}{c}{Input shape} & & \multicolumn{2}{c}{ImageNet-100 \(256^2\)} \\ \cmidrule(lr){1-3} \cmidrule(lr){4-6}
     $Z^{\dagger}$  & $Z^{\star}$  & $Z$  &  & Train PSNR $\uparrow$ & Test PSNR $\uparrow$ \\ \midrule
    $1 \times 1 \times 256$ & $8 \times 8 \times 256$ & $16 \times 16 \times 256$ & & 33.93 & 31.05 \\ 
    $1 \times 1 \times 512$ & $8 \times 8 \times 256$ & $16 \times 16 \times 128$ & & 31.28 & 29.83 \\ 
    \bottomrule
    \end{tabular}
    }
\label{tab:imagenet}
\vspace{-0.25em}
\end{table}
\begin{table}[!t]
    \centering
    \vspace{-0.5em}
    \caption{Image classification performance on CIFAR-10. The table reports the Top-1 accuracy (in \%) along with reconstruction performance across different numbers of augmentations. $\star$ denotes best classification performed results from Spatial Functa~\cite{bauer2023spatial}. $\star\star$ denotes that they used more sophisticated augmentations, including MixUp~\cite{zhang2018mixup} and CutMix~\cite{yun2019cutmix}.}
    \vspace{-0.5em}
    \resizebox{1\linewidth}{!}{%
    \begin{tabular}{l|cc|ccc}
    \toprule
    Method & Test PSNR $\uparrow$ & \# GFLOPs $\downarrow$ & Top-1 Acc ($\uparrow$) & \# Augments & \# Params (M) \\
     \midrule
     ResNet-50~\cite{he2016deep} & -- & -- & \cellcolor{light}95.36 & $\star\star$ & 23.52M \\   
     ResNeXT~\cite{xie2017aggregated} & -- & -- & 94.91 & $\star\star$ & 9.13M \\   
     ViT/4~\cite{dosovitskiy2020image}& -- & -- & 90.61 & $\star\star$ & 6.27M \\   
     Swin/2~\cite{liu2022swin} & -- & -- & 91.10 & $\star\star$ & 7.05 \\   
    \midrule

    $\star$ Functa~\cite{dupont2022data}
     & 31.90 & 3.503 & 68.30 & 50  & 1.58 \\ 

    $\star$ Spatial Functa~\cite{bauer2023spatial} 
     & \cellcolor{purple}37.20 & 0.271\cellcolor{light} & 90.30 & 50  & 7.10 \\ 

    \midrule
    
    \multirow{3}{*}{\textbf{LIFT}} 
     &  & & 91.04 & 1  & 4.04 \\ 
     & 35.95\cellcolor{light} & 0.136\cellcolor{purple} & 95.30 & 5  & 4.04 \\ 
     & & & \cellcolor{purple}95.47 & 10 & 4.04 \\ 

    \bottomrule
    \end{tabular}
    }
\label{tab:classification}
\vspace{-1.25em}
\end{table}

\noindent\textbf{Generation:} We compare LIFT with domain-agnostic generative models (Functa~\cite{dupont2022data}, GEM~\cite{du2021learning}, GASP~\cite{dupont2022generative}, DPF~\cite{zhuang2023diffusion}, and mNIF~\cite{you2024generative}) all aimed at unified generative modeling across various data domains (see~\autoref{tab:2d_img}). In CelebA-HQ, LIFT achieves an FID of 10.0, outperforming DPF and mNIF (both at 13.2), indicating that LIFT produces images with superior fidelity compared to competing methods. While GASP’s 13.5 FID is relatively strong, the generated samples in ~\autoref{fig:gasp_generation} reveal artifacts typical of adversarial approaches. LIFT attains a Precision of 0.900, which is competitive with mNIF-L's 0.902, indicating that the generated images are high quality and closely resemble real data distribution. LIFT also achieves a Recall of 0.631, substantially higher than mNIF-L's 0.544 and DPF's 0.347. This higher recall suggests that LIFT captures a broader diversity of the data distribution, generating a wider variety of plausible images. The F1 score of 0.742 further reflects the model's ability to balance quality and diversity effectively. \autoref{fig:2d_image} provides qualitative examples, highlighting LIFT's ability to generate realistic, artifact-free images.

\noindent\textbf{Classification:} We train LIFT on CIFAR-10~\cite{krizhevsky2010cifar} for 300K iterations, followed by latent modulation generation. Training utilizes a customized VMamba~\cite{liu2024vmamba} model with standard augmentations, including horizontal flips and random cropping, where $32 \times 32$ crops are extracted from a padded $40 \times 40$ image. As shown in ~\autoref{tab:classification}, LIFT achieves competitive performance, reaching 95.47\% Top-1 accuracy with 10 augmentations, surpassing all baselines, including ResNet-50 (95.36\%) and ResNeXT (94.91\%), while using significantly fewer parameters. Increasing the number of augmentations consistently improves classification accuracy. Compared to the global-based method Functa and the spatial-based approach Spatial Functa, LIFT achieves significantly higher accuracy, indicating that global representations struggle to capture task-specific features effectively, while local information alone is insufficient for downstream tasks. In contrast, our method integrates multiscale information, retaining global context while enriching fine-grained details. This hierarchical approach balances scalability and reconstruction quality, outperforming spatial and global methods while using $5\times$ less augmented data.

\vspace{-0.25em}
\subsection{3D Analysis}
\vspace{-0.5em}
\noindent\textbf{Reconstruction:} As shown in~\autoref{tab:2d_img}, LIFT achieves outstanding results with an MSE of 0.00053 and a PSNR of 35.2 dB, setting a new benchmark in reconstruction quality. LIFT notably surpasses the second-best models by a significant margin—achieving an MSE improvement of 28 times and a PSNR boost of 13.75 dB. These results demonstrate LIFT’s ability to deliver high-quality shape reconstruction with a minimal computational footprint, establishing a clear performance advantage over existing models and affirming its effectiveness in precise, efficient modeling. Qualitative results are shown in~\autoref{fig:3d_image}.

\noindent\textbf{Generation:}  We present the 3D voxel generation results in ~\autoref{fig:3d_image} and~\autoref{tab:2d_img}. LIFT achieves comparable performance in both MMD and coverage scores. These results confirm LIFT's ability to generate diverse and accurate shapes that effectively capture the underlying data distribution. Moreover, LIFT is highly resource-efficient, requiring 5× and 47× fewer FLOPs than mNIF-L and GEM, respectively, while achieving comparable generation results and significantly higher reconstruction quality. \autoref{fig:voxel_generation} provides additional visualizations of the generated 3D voxel shapes, further illustrating our model's capabilities.

\vspace{-0.3em}
\subsection{Latent Interpolation}
\vspace{-0.25em}
To assess the quality and structure of LIFT's learned latent space across both 2D and 3D data, we conducted interpolation experiments using challenging input pairs with diverse attributes. For 2D images, as shown in \autoref{fig:2d_interpolation}, each row presents an interpolation between two distinct faces differing in complex features such as gender, age, expressions, hairstyles, and accessories like hats. These challenging scenarios were selected to test the model's ability to handle significant variations in appearance. As the interpolation factor moves from 0 to 1, we observe smooth transitions blending distinct attributes naturally, without introducing artifacts, demonstrating that LIFT’s latent space is well-structured, preserving realism and a smooth manifold. For 3D data, similar interpolations were conducted between distinct shapes with varied structural characteristics (see \autoref{fig:3d_interpolation}). LIFT’s latent space effectively captures complex spatial transformations, enabling smooth transitions between shapes while maintaining structural integrity. 

\vspace{-0.5em}
\subsection{Model Efficiency}
\vspace{-0.25em}
An important aspect of LIFT is its computational efficiency. With only 0.915M learnable parameters and requiring just 54.52M FLOPs (\autoref{tab:2d_img}), LIFT outperforms significantly larger models in both reconstruction and generation tasks. For instance, while mNIF-L comprises 33.4M parameters, it achieves lower performance. Even compared to mNIF-S, the closest efficient model, LIFT requires 1.27× fewer FLOPs. Compared to Functa, LIFT uses 158× fewer FLOPs and 3.6× fewer parameters; compared to GEM, it requires 60.51× fewer FLOPs and 108.20× fewer parameters. This advantage translates to faster encoding in the context adaptation stage and faster decoding for downstream tasks such as generation. LIFT’s design—focused on compact and expressive latent representations through a localized, parallel neural implicit functions framework—enables it to capture essential information with minimal parameters, reducing computational costs without sacrificing performance. Even in 3D, LIFT maintains efficiency with only 8.8M learnable parameters and a FLOP count of just 4.38G. In comparison, GEM, with 99M parameters, demands significantly more computation (207G FLOPs) but achieves lower accuracy. Similarly, while mNIF-S and mNIF-L use fewer parameters, they still fall short in performance metrics.
\section{Ablation Studies}
\noindent\textbf{Different Configurations of LIFT:}
This ablation study on the CelebA-HQ dataset explores how adjustments to the spatial size and channel capacity of three latent variables—\( Z \) (local), \( Z^\star \) (intermediate), and \( Z^\dagger \) (global)—affect Test PSNR and rFID (\autoref{tab:abl}). Increasing the spatial dimensions of \( Z \) from \( 4 \times 4 \) to \( 8 \times 8 \) (rows 6 and 5) improves PSNR from 30.27 to 40.42 and reduces rFID from 21.93 to 2.47, enhancing spatial fidelity by enlarging spatial dimension. Expanding \( Z^\star \)’s spatial size from \( 2 \times 2 \) to \( 4 \times 4 \) yields a smaller PSNR boost (40.42 to 40.91) and rFID drop (2.47 to 2.22), showing that mid-level features provide fewer spatial details than local latents. Doubling channels across all latents (rows 3 and 4) improves PSNR by 8.95 and reduces rFID by 1.82, highlighting that higher channel capacity captures richer representations.
\begin{table}[!t]
    \centering
    \caption{Impact of spatial and channel size in local, intermediate, and global latents on PSNR and rFID.}
    \label{tab:abl}
     \vspace{-0.75em}
    \resizebox{0.8\linewidth}{!}{%
    \begin{tabular}{lllccc}
    \toprule
     \multicolumn{3}{c}{Input shape} & \multirow{2}{*}{Test PSNR $\uparrow$} & \multirow{2}{*}{rFID $\downarrow$} \\ \cmidrule(lr){1-3}
     $Z^{\dagger}$ & $Z^{\star}$ & $Z$ &   \\ \midrule
     $1 \times 1 \times 64$ & $4 \times 4 \times 32$ & $8 \times 8 \times 16$ & 29.00 & 23.51 \\
     $1 \times 1 \times 128$ & $4 \times 4 \times 64$ & $8 \times 8 \times 32$ & 34.38 & 7.87 \\
     $1 \times 1 \times 256$ & $4 \times 4 \times 128$ & $8 \times 8 \times 64$ & 40.91 & 2.22 \\
     $1 \times 1 \times 512$ & $4 \times 4 \times 256$ & $8 \times 8 \times 128$ & 49.86 &  0.40\\
     $1 \times 1 \times 256$ & $2 \times 2 \times 128$ & $8 \times 8 \times 64$ & 40.42 & 2.47 \\
     $1 \times 1 \times 256$ & $2 \times 2 \times 128$ & $4 \times 4 \times 64$ & 30.27 & 21.93 \\
     $1 \times 1 \times 256$ & $4 \times 4 \times 128$ & $8 \times 8 \times 128$ & 49.75 & 0.58 \\
     $1 \times 1 \times 64$ & $4 \times 4 \times 64$ & $8 \times 8 \times 64$ & 40.72 & 2.38 \\
     $1 \times 1 \times 128$ & $4 \times 4 \times 128$ & $8 \times 8 \times 128$ & 49.61 & 0.62 \\ 
     \bottomrule
    \end{tabular}}
    \vspace{-1em}
\end{table}

\noindent\textbf{Impact of Hierarchical Latent Generator:} To analyze the contribution of HLG in our model, we compare the performance of LIFT with and without HLG over 200K training iterations, as shown in ~\autoref{tab:hlg_contribution}. Without HLG, the model relies solely on local latents without multiscale encoding, leading to a performance drop. Specifically, the Test PSNR after 200K iterations decreases from 30.91 to 39.28 (a 4.15\% reduction), indicating lower reconstruction quality, while rFID increases from 2.22 to 2.70 (a 17.8\% rise), suggesting weaker alignment with the real data distribution. These results highlight the importance of our hierarchical design in effectively incorporating multiscale information.
\begin{table}[t]
    \centering
    \caption{Performance comparison of LIFT with and without HLG across different training iterations.}
    \vspace{-0.5em}
    \resizebox{0.95\linewidth}{!}{%
    \begin{tabular}{@{}lcccccc@{}} 
        \toprule
        Model & \multicolumn{4}{c}{Training PSNR $\uparrow$} & Test PSNR $\uparrow$ & rFID $\downarrow$ \\
        \cmidrule(lr){2-5}
        & 50K & 100K & 150K & 200K & & \\
        \midrule
        LIFT & 39.31 & 40.57 & 40.97 & 41.12 & 40.91 & 2.22 \\
        LIFT w/o HLG & 38.94 \scriptsize{(-1.0\%)} & 39.43 \scriptsize{(-2.8\%)} & 39.56 \scriptsize{(-3.4\%)} & 40.08 \scriptsize{(-2.5\%)} & 39.28 \scriptsize{(-4.15\%)} & 2.70 \scriptsize{(+17.8\%)} \\
        \bottomrule
    \end{tabular}
    \label{tab:hlg_contribution}
    }
    \vspace{-1em}
\end{table}
\noindent\textbf{Impact of Local P-MLP Architecture on Reconstruction and Efficiency:} We evaluate the impact of depth and the number of hidden layers in our local P-MLP for image and shape reconstruction tasks using the same HLG configurations (\autoref{tab:PMLP_depth}). With just 100K training steps, our model surpasses mNIF-L while using 17.18 times fewer FLOPs. Interestingly, for 3D data, we achieve results close to our benchmark LIFT while using 4 times fewer FLOPs.
\begin{table}[!t]
    \centering
    \caption{Impact of local P-MLP depth and width on reconstruction quality and computational efficiency, with 100K training iterations for images and 120K for shapes.}
    \label{tab:PMLP_depth}
    \vspace{-0.5em}
    \resizebox{0.9\linewidth}{!}{%
    \begin{tabular}{cccclll}
    \toprule
    \multicolumn{7}{c}{CelebA-HQ} \\ \midrule
    \multirow{2}{*}{\# Hidden Layers} & \multirow{2}{*}{Width} & \multirow{2}{*}{Test PSNR $\uparrow$} & \multirow{2}{*}{rFID / MSE $\downarrow$} & \multicolumn{2}{c}{\# Params} & \multirow{2}{*}{\# FLOPs} \\ \cmidrule(lr){5-6} 
     &  &  &  & Training & Inference \\ \midrule
    1 & 256 & 40.25 & 2.36 & 4.417M & 4.348M & 277.34M  \\
    4 & 64 & 39.09 & 2.78 & 1.185M & 1.110M & 71.30M  \\
    8 & 32 & 38.53 & 3.21 & 0.646M & 0.572M & 36.96M  \\
    16 & 16 & 35.16 & 5.91 & 0.377M & 0.403M & 19.79M  \\ \cmidrule(lr){1-7} 
    \multicolumn{7}{c}{ShapeNet} \\ \cmidrule(lr){1-7} 
    4 & 64 & 35.15 & 0.00053 & 8.828M & 8.185M & 4.38G  \\
    8 & 32 & 34.28 & 0.00055 & 4.548M & 4.445M & 1.12G  \\ \bottomrule
    \end{tabular}
    }
\end{table}

\noindent\textbf{Impact of $T_{\text{inner}}$:} We evaluate the impact of inner loop gradient steps on test performance after meta-learning, as shown in \autoref{tab:T_inner}. With $T_{\text{inner}} = 1$, the model achieves limited performance (PSNR of 33.25, rFID of 7.60). Increasing to 2 steps significantly boosts PSNR to 39.56 and reduces rFID to 2.55. At $T_{\text{inner}} = 3$, performance further improves to a PSNR of 40.17 and rFID of 2.22, making it a reliable choice. Additional steps ($T_{\text{inner}} = 4$) yield only marginal gains, indicating most benefits are realized by 2–3 steps.
\begin{table}[!t]
    \centering
    \vspace{-0.5em}
    \caption{Impact of $T_{\text{inner}}$ step on Test PSNR and rFID.}
    \vspace{-0.5em}
    \resizebox{0.6\linewidth}{!}{%
    \begin{tabular}{l|cccc}
    \toprule
    $T_\text{inner}$ & 1 & 2 & 3 & 4 \\ \midrule
    Test PSNR $\uparrow$ & 33.25 & 39.56 & 40.17 & 40.28 \\
    rFID $\downarrow$ & 7.60 & 2.55 & 2.22 & 2.15 \\
    \midrule
    \end{tabular}
    }
    \label{tab:T_inner}
\vspace{-1.25em}
\end{table}

\noindent\textbf{Ablation on ReLIFT:} We integrate ReLIFT into LIFT by modifying SIREN-based modulation and evaluate it on CelebA-HQ reconstruction. With $\gamma=2$ and 120K iterations, ReLIFT accelerates convergence, achieving 2.3 higher PSNR than LIFT at 10K iterations and 38.21 at 20K, outperforming all counterparts. It reaches 10K in 68 min and 20K in 134 min (batch size 128, 4 A40 GPUs), making it efficient for rapid, high-quality reconstructions. Although ReLIFT achieves a higher early PSNR, LIFT ultimately surpasses it, as higher training is necessary for downstream tasks~\cite{you2024generative}. We use ReLIFT for single-task signal representation and inverse problems, enhancing SIREN’s expressivity without extra parameters and bridging the convergence-capacity gap in SOTA models (see SM for details).
\vspace{-0.25em}

\section{Conclusion}
We introduced LIFT, a task-agnostic framework that captures multiscale data features using meta-learning and a hierarchical latent generator, delivering efficient, high-fidelity representations across data modalities. LIFT, along with ReLIFT, which adds residual connections and frequency scaling, achieves SOTA results in generative modeling, classification, and single data analysis, with notable computational efficiency. Our framework sets a new benchmark for scalable implicit representations, offering a versatile and powerful approach for unified task modeling. 

\section{Appendix}
\appendix

\section{More of LIFT}

\subsection{Qualitative Comparison: LIFT vs. GASP}
We generated 10 samples from GASP~\cite{dupont2022generative} using the pre-trained model available on their GitHub repository. Although GASP achieves a competitive FID score of 13.5, its adversarial approach often leads to noticeable artifacts in the generated samples, a common issue with adversarial methods. These artifacts suggest limitations in capturing fine-grained details and spatial coherence. In contrast, our model’s samples demonstrate greater visual fidelity, with fewer artifacts and a smoother, more coherent structure across generated outputs as shown in Figure 2 in the paper.
\begin{figure}[h]
    \centering
    \includegraphics[width=\linewidth]{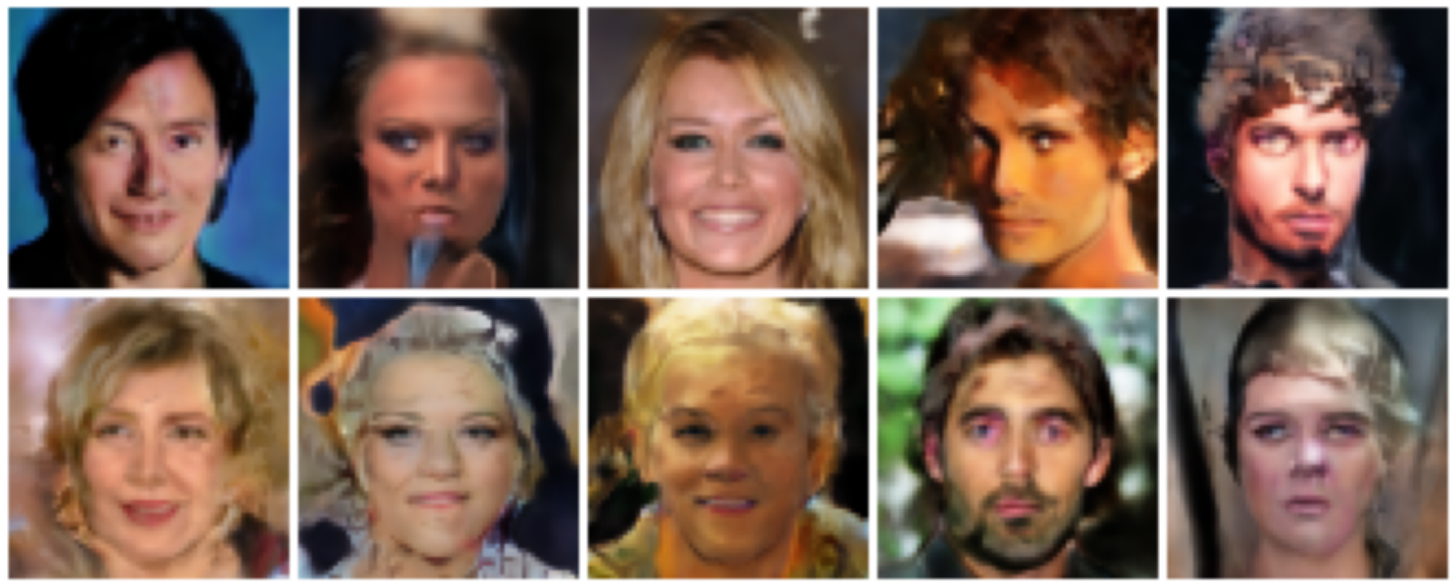}\\
    \caption{Generation result of GASP~\cite{dupont2022generative}.}
    \vspace{-0.5em}
    \label{fig:gasp_generation}
\end{figure}
\begin{figure*}[h]
    \centering
    \includegraphics[width=\linewidth]{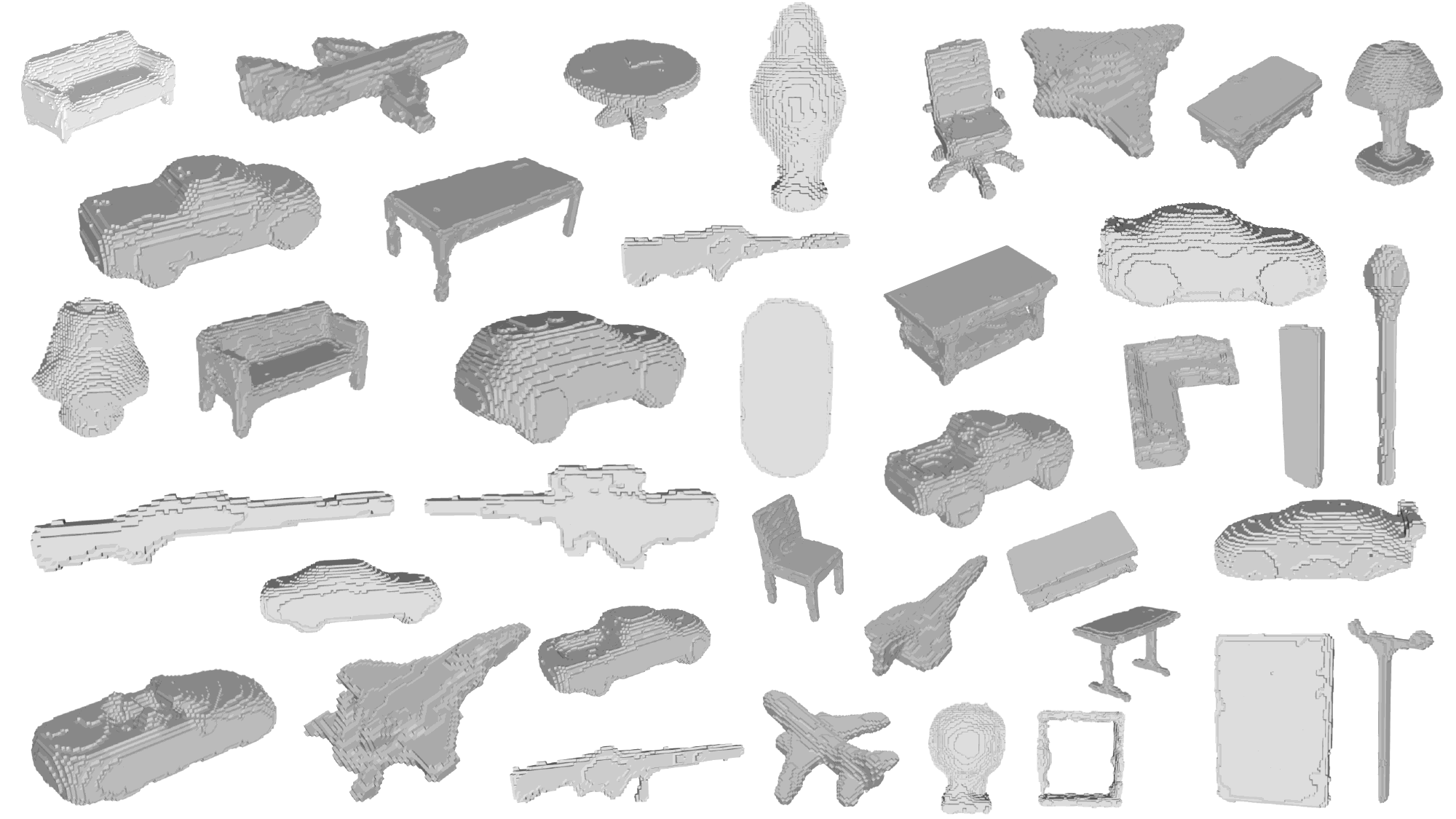}\\
    \caption{LIFT Shapenet voxel generation results.}
    \vspace{-0.5em}
    \label{fig:voxel_generation}
\end{figure*}

\subsection{Ablation on scale levels of HLG}
To assess the contribution of each scale in our HLG framework, we conduct an ablation study on the CelebA-HQ dataset by selectively removing individual scales and measuring the resulting performance. Specifically, we evaluate four configurations: (1) Global only, (2) Local only, (3) Global + Local, and (4) Global + Intermediate + Local (our full model). \autoref{abl:scale} summarizes the corresponding PSNR values on the test set. The Global-only configuration yields a PSNR of 24.82 dB, underscoring the importance of fine-scale information. Conversely, using the Local scale alone significantly improves performance to 39.28 dB, highlighting the role of detailed spatial features. Combining the Global and Local scales further raises the PSNR to 39.62 dB, indicating that multi-scale cues complement each other. Finally, incorporating the Intermediate scale and the Global and Local scales produces the highest PSNR of 40.91 dB, demonstrating that each scale contributes unique and essential information. These results confirm that a multi-scale approach is crucial for capturing both coarse global context and finer local details, thereby improving overall reconstruction quality.
\begin{table}[!th]
    \centering
     \caption{Comparison of test PSNR across different scales.}
    \resizebox{\linewidth}{!}{
    \begin{tabular}{c|cccc}
    \toprule
    \textbf{Scale}          & \textbf{Global} & \textbf{Local} & \textbf{Global-Local} & \textbf{Global-Intermediate-Local} \\ \midrule
    \textbf{Test PSNR} & 24.82           & 39.28          & 39.62                     & 40.91                              \\ \bottomrule
    \end{tabular}
    }
    \label{abl:scale}
\end{table}

\subsection{Different Configurations of LIFT on the ShapeNet dataset}
In~\autoref{tab:abl_imp}, we evaluate the impact of channel sizes in the latent representations on reconstruction quality using PSNR for the ShapeNet \(64^3\) dataset. Our main configuration employs a global latent \(Z^{\dagger}\) with 256 channels, an intermediate latent \(Z^{\star}\) with 128 channels, and a local latent \(Z\) with 64 channels, achieving a PSNR of 35.15 dB with a compression ratio (CR) of 8× for $Z^{\alpha}$. Halving the channel dimensions leads to a PSNR of 33.91 dB (CR of 16×). Further reduction to 64, 32, and 16 channels yields a PSNR of 32.21 dB (CR of 32×). Additional configurations with even smaller channel sizes—32, 16, 8 channels and 32, 16, 2 channels—result in PSNR values of 30.78 dB (CR of 64×) and 24.01 dB (CR of 256×), respectively. These experiments clearly demonstrate the trade-off between compression and reconstruction fidelity that higher channel capacities in the latent spaces contribute to better reconstruction fidelity, while more compressed representations, despite reducing PSNR, can still maintain competitive performance relative to other methods.

\begin{table}[!th]
    \centering
    \caption{Impact of channel size in local, intermediate, and global latents on PSNR. CR denotes the compression ratio based on an input shape of $64^3$. Higher CR values indicate greater compression.}
    \label{tab:abl_imp}
    \resizebox{1\linewidth}{!}{%
    \begin{tabular}{lllccc}
    \toprule
     \multicolumn{3}{c}{Input shape} & \multirow{2}{*}{Test PSNR $\uparrow$} & \multirow{2}{*}{CR} \\ \cmidrule(lr){1-3}
     $Z^{\dagger}$ & $Z^{\star}$ & $Z$ &   &  \\ \midrule
     $1 \times 1 \times 1 \times 256$ & $4 \times 4 \times 4 \times 128$ & $8 \times 8 \times 8 \times 64$ & 35.15 & 8$\times$ \\
     $1 \times 1 \times 1 \times 128$ & $4 \times 4 \times 4 \times 64$ & $8 \times 8 \times 8 \times 32$ & 33.91 & 16$\times$  \\
     $1 \times 1 \times 1 \times 64$ & $4 \times 4 \times 4 \times 32$ & $8 \times 8 \times 8 \times 16$ & 32.21 & 32$\times$  \\
     $1 \times 1 \times 1 \times 32$ & $4 \times 4 \times 4 \times 16$ & $8 \times 8 \times 8 \times 8$ & 30.78 & 64$\times$  \\
     $1 \times 1 \times 1 \times 32$ & $4 \times 4 \times 4 \times 16$ & $8 \times 8 \times 8 \times 2$ & 24.01 & 256$\times$  \\
     \bottomrule
    \end{tabular}}
    \vspace{-1em}
\end{table}

\section{Implementation Details}
\begin{table*}[t]
\caption{Training configurations and model specifications for Stage 1 across CelebA-HQ, ImageNet, and ShapeNet datasets, detailing hyperparameters optimized for each dataset.}
\resizebox{\textwidth}{!}{%
\begin{tabular}{lllccccccccccc}
\toprule
\multicolumn{14}{c}{CelebA-HQ $64^2$} \\ \midrule
\multicolumn{3}{c}{Input shape} & \multirow{2}{*}{} & \multicolumn{3}{c}{P-MLP Configs} & \multirow{2}{*}{} & \multicolumn{6}{c}{Meta-learning Configs} \\ \cmidrule(lr){1-3} \cmidrule(lr){5-7} \cmidrule(lr){9-14} 
$Z^{\dagger}$ & $Z^{\star}$ & $Z$ &  & \# Hidden Layers & Width & $W_0$ &  & Inner lr & Inner Opt & Outer lr & Outer Opt & $T_{\text{inner}}$ & Meta-SGD \\ \midrule
$1 \times 1 \times 64$ & $4 \times 4 \times 32$ & $8 \times 8 \times 16$ &  & 1 & 256 & 15 &  & 1.0 & SGD & 1.5e-5 & Adam & 3 & \ding{51} \\
$1 \times 1 \times 128$ & $4 \times 4 \times 64$ & $8 \times 8 \times 32$ &  & 1 & 256 & 20 &  & 1.0 & SGD & 5e-5 & Adam & 3 & \ding{51} \\
$1 \times 1 \times 256$ & $4 \times 4 \times 128$ & $8 \times 8 \times 64$ &  & 1 & 256 & 20 &  & 1.0 & SGD & 5e-5 & Adam & 3 & \ding{51} \\
$1 \times 1 \times 512$ & $4 \times 4 \times 256$ & $8 \times 8 \times 128$ &  & 1 & 256 & 20 &  & 1.0 & SGD & 3e-5 & Adam & 3 & \ding{51} \\ \rowcolor{pink} 
$1 \times 1 \times 256$ & $4 \times 4 \times 128$ & $8 \times 8 \times 64$ &  & 1 & 256 & 20 &  & 1.0 & SGD & 5e-5 & Adam & 3 & \ding{51} \\ \rowcolor{orange} 
$1 \times 1 \times 256$ & $4 \times 4 \times 128$ & $8 \times 8 \times 64$ & 
\multicolumn{1}{l}{} & 3 & 64 & 20 &  & 1.0 & SGD & 5e-5 & Adam & 3 & \ding{51} \\

$1 \times 1 \times 256$ & $4 \times 4 \times 128$ & $8 \times 8 \times 64$ & 
\multicolumn{1}{l}{} & 4 & 64 & 20 &  & 1.0 & SGD & 5e-5 & Adam & 3 & \ding{51} \\
$1 \times 1 \times 256$ & $4 \times 4 \times 128$ & $8 \times 8 \times 64$ & 
\multicolumn{1}{l}{} & 8 & 32 & 20 &  & 1.0 & SGD & 3e-5 & Adam & 3 & \ding{51} \\
$1 \times 1 \times 256$ & $4 \times 4 \times 128$ & $8 \times 8 \times 64$ & 
\multicolumn{1}{l}{} & 16 & 16 & 20 &  & 1.0 & SGD & 2e-5 & Adam & 3 & \ding{51} \\

$1 \times 1 \times 256$ & $2 \times 2 \times 128$ & $8 \times 8 \times 64$ &  & 1 & 256 & 20 &  & 1.0 & SGD & 5e-5 & Adam & 3 & \ding{51} \\
$1 \times 1 \times 256$ & $2 \times 2 \times 128$ & $4 \times 4 \times 64$ &  & 1 & 256 & 20 &  & 1.0 & SGD & 5e-5 & Adam & 3 & \ding{51} \\
$1 \times 1 \times 256$ & $4 \times 4 \times 128$ & $8 \times 8 \times 128$ &  & 1 & 256 & 20 &  & 1.0 & SGD & 4e-5 & Adam & 3 & \ding{51} \\
$1 \times 1 \times 64$ & $4 \times 4 \times 64$ & $8 \times 8 \times 64$ &  & 1 & 256 & 20 &  & 1.0 & SGD & 5e-5 & Adam & 3 & \ding{51} \\
$1 \times 1 \times 128$ & $4 \times 4 \times 128$ & $8 \times 8 \times 128$ &  & 1 & 256 & 20 &  & 1.0 & SGD & 4e-5 & Adam & 3 & \ding{51} \\ \midrule
\multicolumn{14}{c}{ImageNet-100 $256^2$} \\ \midrule
$1 \times 1 \times 256$ & $8 \times 8 \times 256$ & $16 \times 16 \times 256$ & \multicolumn{1}{l}{} & 2 & 256 & 20 &  & 1.0 & SGD & 2e-5 & Adam & 3 & \ding{51} \\ 
$1 \times 1 \times 512$ & $8 \times 8 \times 256$ & $16 \times 16 \times 128$ & \multicolumn{1}{l}{} & 2 & 256 & 20 &  & 1.0 & SGD & 2e-5 & Adam & 3 & \ding{51} \\ \midrule
\multicolumn{14}{c}{CIFAR-10 $32^2$} \\ \midrule
$1 \times 1 \times 64$ & $4 \times 4\times 32$ & $8 \times 8 \times 16$ & \multicolumn{1}{l}{} & 2 & 256 & 10 &  & 1.0 & SGD & 3e-5 & Adam & 3 & \ding{51} \\  \midrule
\multicolumn{14}{c}{ShapeNet $64^3$} \\ \midrule
$1 \times 1 \times 1 \times 256$ & $4 \times 4 \times 4 \times 128$ & $8 \times 8 \times 8 \times 64$ & \multicolumn{1}{l}{} & 4 & 64 & 20 &  & 1.0 & SGD & 4e-5 & Adam & 3 & \ding{51} \\
$1 \times 1 \times 1 \times 256$ & $4 \times 4 \times 4 \times 128$ & $8 \times 8 \times 8 \times 64$ & \multicolumn{1}{l}{} & 8 & 32 & 20 &  & 1.0 & SGD & 4e-5 & Adam & 3 & \ding{51}\\
$1 \times 1 \times 1 \times 128$ & $4 \times 4 \times 4 \times 64$ & $8 \times 8 \times 8 \times 32$ & \multicolumn{1}{l}{} & 4 & 64 & 20 &  & 1.0 & SGD & 4e-5 & Adam & 3 & \ding{51}\\
$1 \times 1 \times 1 \times 64$ & $4 \times 4 \times 4 \times 32$ & $8 \times 8 \times 8 \times 16$ & \multicolumn{1}{l}{} & 4 & 64 & 20 &  & 1.0 & SGD & 4e-5 & Adam & 3 & \ding{51}\\
$1 \times 1 \times 1 \times 32$ & $4 \times 4 \times 4 \times 16$ & $8 \times 8 \times 8 \times 8$ & \multicolumn{1}{l}{} & 4 & 64 & 20 &  & 1.0 & SGD & 4e-5 & Adam & 3 & \ding{51}\\
$1 \times 1 \times 1 \times 32$ & $4 \times 4 \times 4 \times 16$ & $8 \times 8 \times 8 \times 2$ & \multicolumn{1}{l}{} & 4 & 64 & 20 &  & 1.0 & SGD & 2e-5 & Adam & 3 & \ding{51}\\
\bottomrule
\end{tabular}
}
\label{tab:configs}
\end{table*}

\subsection{Datasets}
\subsubsection{Images}
We use the CelebA-HQ \(64 \times 64\) dataset~\cite{karras2017progressive}, partitioned into 27K training and 3K testing images, as provided by Functa~\cite{dupont2022data}. For reconstruction tasks, performance is evaluated using the Peak Signal-to-Noise Ratio (PSNR) and reconstruction Fréchet Inception Distance (rFID). Additionally, the FID metric~\cite{heusel2017gans} is computed alongside precision, recall, and F1 score metrics to assess image generation quality~\cite{sajjadi2018assessing, naeem2020reliable}. To further demonstrate scalability, we use the ImageNet-100 \(256 \times 256\) dataset~\cite{russakovsky2015imagenet} for the reconstruction task. ImageNet-100~\cite{russakovsky2015imagenet,imagenet100_kaggle}, a subset of the ImageNet-1k dataset, consists of 100 classes with 130K training and 5K test samples. For the classification task, we utilize the CIFAR-10 dataset~\cite{cifar10}, which contains 60K \(32 \times 32\) color images distributed across 10 classes, with 6K images per class. The dataset is divided into 50K training images and 10K test images.

\subsubsection{Voxels}
We employ the ShapeNet dataset \cite{chang2015shapenet}, which includes 35,019 training and 8,762 testing samples for 3D voxel data. Each voxel is represented at a \(64^3\) resolution with 16,384 surface points. We evaluate reconstruction tasks using Mean Squared Error (MSE) and PSNR metrics. For generation tasks, following IM-Net \cite{chen2019learning}, we generate 8,762 shapes and extract 2,048-dimensional mesh features. Performance is measured using coverage, maximum mean discrepancy (MMD) \cite{achlioptas2018learning}, and Chamfer distance, adhering to the GEM~\cite{du2021learning} and mNIF~\cite{you2024generative} protocols.

\subsection{Stage 1: Context Adaptation}
For stage 1, we implement all models in JAX \citep{bradbury2018jax} using Haiku \citep{haiku2020github} and JAXline for training, with Functa \cite{dupont2022data} as the baseline framework for our coding setup. This stage involves adapting models for image and voxel data through a meta-learning framework, as detailed in the subsections below. 

\subsubsection{Images}
\label{imp:context:images}
We train on images from the train split of the CIFAR-10 \(32^2\), CelebA-HQ \(64^2\), and ImageNet-100 \(256^2\) datasets. The model takes local 2D input coordinates \((x_{\text{local}}, y_{\text{local}})\), with each P-MLP processing its corresponding local coordinate and returning 3D RGB values \((y_R, y_G, y_B)\) representing the local patch. After processing, we merge these patches to reconstruct the full image. We use local dense sampling, querying all local input coordinates in parallel for all P-MLPs. For instance, with a grid size of 8, we generate 64 query coordinates \((64/8 \times 64/8)\) for $8 \times 8$ P-MLPs to input into the network for image benchmarking. In addition, the model is trained using a meta-learning approach for 200K iterations with a per-device batch size of 128 on four A40 GPUs (each with 48 GB of memory) using the CelebA-HQ dataset. We used a per-device batch size of 512 for CIFAR-10 and 16 for ImageNet-100, trained on 4 A40 GPUs. All details regarding different model configurations and meta-learning setups are provided in \autoref{tab:configs}. To create the modulation datasets during inference, we freeze the network weights and optimize the zero-initialized latents in 3 steps using SGD. We then save the resulting \(Z^\alpha\) latent vectors for both training and test data. Additionally, we set \( K = 8 \) in our \(\mathcal{L_{\text{smoothness}}}\) loss function with the $\lambda=(1/100)^2$. For the ablation studies, we use the highlighted \colorbox{pink}{pink} in~\autoref{tab:configs}. Our results in the main Table 1 use the highlighted \colorbox{orange}{orange}.

\subsubsection{Voxels}
We train on the ShapeNet \(64^3\) dataset, using 35,019 training shapes at \(64 \times 64 \times 64\) resolution and testing on 8,762 shapes at the same resolution. The model takes local 3D input coordinates \((x_{\text{local}}, y_{\text{local}}, z_{\text{local}})\), with each P-MLP processing its corresponding local coordinate, and returns a scalar output \(y_o\), indicating whether the queried coordinate is inside or outside of the object. After processing, we merge these local outputs to reconstruct the entire voxel grid representation of the shape. The model is trained using a meta-learning approach, configured for 120K iterations with a per-device batch size of 8, on four A40 GPUs. Further details on the configurations and meta-learning setups are provided in~\autoref{tab:configs}. We use the approach described for \textbf{Images} (~\autoref{imp:context:images}) to create the modulation datasets during inference. We set \( K = 16 \) in our \(\mathcal{L_{\text{smoothness}}}\) loss function with the $\lambda=(1/100)^2$.

\subsection{Stage 2: Task-Driven Generalization}
\subsubsection{Images}
\noindent\textbf{Generation:} We base our image generation implementation on the ablated diffusion model (ADM)~\cite{dhariwal2021diffusion}. The training pipeline begins with the creation of our modulation dataset using the LIFT framework. In this setup, the latent representations of \( Z^{\alpha} \) are derived using the configurations highlighted in  \colorbox{orange}{orange} in Table~\ref{tab:configs}.
For the diffusion model, we configure an image size of 8 and set the number of channels to 320, employing channel multipliers of 1, 2, and 4. Each layer consists of two residual blocks with a dropout rate of 0.1. Attention mechanisms are applied at a resolution of 4, and both \texttt{resblock\_updown } and \texttt{use\_scale\_shift\_norm} are enabled to enhance feature normalization and spatial resolution handling. The diffusion process is conducted over 2000 steps for 500K iterations using a cosine noise schedule. Training is performed with a learning rate of \(1 \times 10^{-4}\) and a batch size of 256. We utilize a loss-second-moment schedule sampler to optimize the training dynamics. To mitigate the low variance observed in the learned modulations, we standardize the latent representations across spatial dimensions ($\frac{Z^\alpha - \mu}{\tau. \sigma}$), ensuring that each feature dimension is centered and scaled independently. This normalization simplifies the generative process. Additionally, we apply a scaling factor ($\tau$) of \(2.5\) to our standardization to enhance training stability. During the sampling phase, we employ the DDIM sampler~\cite{song2021denoising} with a timestep respacing of 200.

\noindent\textbf{Classification:} In our implementation, the VMamba~\cite{zhu2024vision} model is configured with a batch size of 512 and an embedding dimension of 128. Label smoothing with a factor of 0.1 is applied to enhance generalization, and a drop path rate of 0.1 is employed for regularization. The model architecture comprises two stages with depths of 9 and 2 layers, respectively. Latents are normalized across examples using a scaling normalization factor of 2.0. The learning rate schedule includes a base learning rate of 5e-3, a minimum learning rate of 5e-6, and a warmup phase of 10 epochs starting at 5e-5. 

\subsubsection{Voxels}
\noindent\textbf{Generation:}
For our base diffusion model architecture, we use ADM~\cite{dhariwal2021diffusion}, which features a U-Net architecture initially created for 2D image synthesis. Since the original design was intended for 2D, we adapted all operations to function in 3D. We trained our diffusion model for 500K iterations with a learning rate of \( 1 \times 10^{-4} \) and a batch size of 64. The training process employs the "loss-second-moment" schedule sampler, and we standardize the latents like images by scaling them with a factor \( \tau = 2.5 \). The model operates on \( 8 \times 8 \times 8 \) image inputs, with an initial channel size of 128 and channel multipliers set to \{1, 2, 4\}. Each resolution includes two residual blocks. A dropout rate of 0.1 is applied. In addition, \texttt{resblock\_updown } and \texttt{use\_scale\_shift\_norm} are enbaled. The diffusion process comprises 1000 steps and follows a cosine noise schedule for smooth and stable noise transitions. For sampling, DDIM is used with 200 timesteps, achieving a balance between computational efficiency and sample quality. 

\section{ReLIFT vs. SIREN}
\subsection{SIREN Pipeline} Given a set of signals \(\{(\mathbf{r}_i, \mathbf{y}_i)\}_{i=1}^N\), where \(\mathbf{r}_i \in \mathbb{R}^d\) represents spatial coordinates and \(\mathbf{y}_i \in \mathbb{R}^m\) denotes the corresponding attributes, SIREN aims to approximate a continuous function \(f(\mathbf{r}; \theta)\) that maps coordinates to their associated values with high fidelity. This function is parameterized as a neural network, where each layer \(l\) operates according to the following equations:
\begin{equation}
\begin{aligned}
    &\mathbf{z}^{(0)} = \sin\left(\omega_0 (W^{(0)} \mathbf{r}+ \mathbf{b}^{(0)})\right) \\
    &\mathbf{z}^{(l)} = \sin\left(\omega_0 (W^{(l)} \mathbf{z}^{(l-1)} + \mathbf{b}^{(l)})\right), \quad l = 1, \dots, L-2, \\
    &f(\mathbf{r}; \theta) = W^{(L)} \mathbf{z}^{(L-1)} + \mathbf{b}^{(L)},
\end{aligned}
\end{equation}
where \(\mathbf{z}^{(l)}\) is the output of the \(l\)-th layer, \(\theta = \{W^{(l)}, \mathbf{b}^{(l)} \mid l = 1, \dots, L\}\) denotes the learnable parameters, \(L\) is the total number of layers, and \(\omega_0\) is a fixed hyperparameter that scales the input to the sinusoidal activation, controlling the frequency response of each layer. Also, they showed that rescaling initialization by \(\omega_0\) adjusts SIRENs' spectral bias, with higher \(\omega_0\) favoring higher frequencies.

\subsection{Revisiting Input Layer Transformation}
\label{sec:revisiting}
As highlighted in~\cite{yuce2022structured}, the initial layer of SIREN serves as a frequency encoding mechanism. Specifically, the output of the first layer can be expressed as:
\begin{equation}
\mathbf{z}^{(0)} = \sin\left(\Omega \mathbf{r}\right),
\end{equation}
where \(\Omega = \omega_0 W^{(0)} \in \mathbb{R}^{T}\). Considering a three-layer SIREN, the network can be defined as:
\begin{equation}
f(\mathbf{r}; \theta) = \mathbf{w}^{(2)\top} \sin\left( \mathbf{W}^{(1)} \sin\left( \Omega \mathbf{r} \right) \right),
\end{equation}
with \(\mathbf{W}^{(1)} \in \mathbb{R}^{F \times T}\) and \(\mathbf{w}^{(2)} \in \mathbb{R}^F\). The input to each neuron in the second layer is a linear combination of sinusoids at frequencies determined by \(\Omega\). The output of a neuron \(z_m^{(1)}\) in the second layer can be written as:
\begin{align}
z_m^{(1)} &= \sin\left( \mathbf{W}_{m:}^{(1)} \sin\left( \Omega \mathbf{r} \right) \right) \nonumber \\
&= \sin\left( \sum_{t=0}^{T-1} W_{m,t}^{(1)} \sin\left( \boldsymbol{\omega}_t^\top \mathbf{r} \right) \right),
\end{align}
where \(\boldsymbol{\omega}_t^\top\) denotes the \(t\)-th row of \(\Omega\). To analyze the frequency components without considering the bias term, we expand the sine of a sum using Bessel function identities. Using the property:
\begin{equation}
\sin\left( \sum_{k} \phi_k \right) = \operatorname{Im}\left\{ \exp\left( j \sum_{k} \phi_k \right) \right\},
\end{equation}
and the exponential identity:
\begin{equation}
\exp\left( j \sum_{k} \phi_k \right) = \prod_{k} \exp\left( j \phi_k \right),
\end{equation}
we can write:
{\small
\begin{align}
\exp\left( j \sum_{t=0}^{T-1} W_{m,t}^{(1)} \sin\left( \boldsymbol{\omega}_t^\top \mathbf{r} \right) \right) &= \prod_{t=0}^{T-1} \exp\left( j W_{m,t}^{(1)} \sin\left( \boldsymbol{\omega}_t^\top \mathbf{r} \right) \right).
\end{align}}
Using the expansion of the complex exponential of a sine function:
\begin{equation}
\exp\left( j \beta \sin(\theta) \right) = \sum_{n=-\infty}^{\infty} J_n(\beta) e^{j n \theta},
\end{equation}
where \( J_n(\beta) \) is the Bessel function of the first kind of order \( n \), we have:
\begin{align}
\exp\left( j W_{m,t}^{(1)} \sin\left( \boldsymbol{\omega}_t^\top \mathbf{r} \right) \right) = \sum_{s_t=-\infty}^{\infty} J_{s_t}\left( W_{m,t}^{(1)} \right) e^{j s_t \boldsymbol{\omega}_t^\top \mathbf{r}}.
\end{align}
Therefore, the product over \( t \) becomes:
\footnotesize
\begin{align}
\exp\left( j \sum_{t=0}^{T-1} W_{m,t}^{(1)} \sin\left( \boldsymbol{\omega}_t^\top \mathbf{r} \right) \right) 
&= \prod_{t=0}^{T-1} \sum_{s_t=-\infty}^{\infty} J_{s_t}\left( W_{m,t}^{(1)} \right) e^{j s_t \boldsymbol{\omega}_t^\top \mathbf{r}} \nonumber \\
&= \sum_{\mathbf{s} \in \mathbb{Z}^T} \left( \prod_{t=0}^{T-1} J_{s_t}\left( W_{m,t}^{(1)} \right) e^{j s_t \boldsymbol{\omega}_t^\top \mathbf{r}} \right),
\end{align}
\normalsize 
where \( \mathbf{s} = (s_0, s_1, \dots, s_{T-1}) \). Since \( \sin\left( \sum_{t} \phi_t \right) = \operatorname{Im}\left\{ \exp\left( j \sum_{t} \phi_t \right) \right\} \), we have:
\begin{align}
z_m^{(1)} &= \operatorname{Im}\left\{ \exp\left( j \sum_{t=0}^{T-1} W_{m,t}^{(1)} \sin\left( \boldsymbol{\omega}_t^\top \mathbf{r} \right) \right) \right\} \nonumber \\
&= \operatorname{Im}\left\{ \sum_{\mathbf{s} \in \mathbb{Z}^T} \left( \prod_{t=0}^{T-1} J_{s_t}\left( W_{m,t}^{(1)} \right) e^{j s_t \boldsymbol{\omega}_t^\top \mathbf{r}} \right) \right\} \nonumber \\
&= \sum_{\mathbf{s} \in \mathbb{Z}^T} \left( \prod_{t=0}^{T-1} J_{s_t}\left( W_{m,t}^{(1)} \right) \right) \sin\left( \sum_{t=0}^{T-1} s_t \boldsymbol{\omega}_t^\top \mathbf{r} \right).
\end{align}
Thus, the output of the network can be expressed as:
\begin{equation}
\begin{aligned}
f(\mathbf{r}; \theta) = &\sum_{m=0}^{F-1} w_m^{(2)} z_m^{(1)} = \sum_{m=0}^{F-1} w_m^{(2)} \\
&\sum_{\mathbf{s} \in \mathbb{Z}^T} \left( \prod_{t=0}^{T-1} J_{s_t}\left( W_{m,t}^{(1)} \right) \right) 
\sin\left( \sum_{t=0}^{T-1} s_t \boldsymbol{\omega}_t^\top \mathbf{r} \right).
\end{aligned}
\end{equation}
This expression indicates that the network output is a sum of sinusoidal functions at frequencies \( \sum_{t=0}^{T-1} s_t \boldsymbol{\omega}_t \), where \( s_t \in \mathbb{Z} \). This implies that by scaling \( \sum_{t=0}^{T-1} s_t \boldsymbol{\omega}_t \), we can increase the network’s capacity to learn higher-frequency components. In addition, the coefficients of these sinusoids are determined by the products of Bessel functions \( J_{s_t}\left( W_{m,t}^{(1)} \right) \) and the weights \( w_m^{(2)} \). Due to the properties of Bessel functions, which generally decrease in magnitude with increasing order \( |s_t| \) when the argument \( W_{m,t}^{(1)} \) is small, higher-order harmonics (those with larger \( |s_t| \)) tend to have smaller coefficients. This results in an implicit bias towards lower-frequency components, concentrating most of the output signal's energy around the fundamental frequencies \( \boldsymbol{\omega}_t \). However, increasing the scale of inner layer coefficients, like \( W^{(1)} \), boosts higher-order harmonics, allowing the network to capture a wider frequency range. Therefore, we begin by examining how scaling the input layer frequency affects the network’s capacity. Next, we introduce a residual connection term, which intuitively directs the network’s focus towards adjusting \( W^{(1)} \) to amplify higher-order harmonics while preserving the fundamental components.

\subsection{Effect of Input Scaling on Frequency Representation}

To investigate the effect of scaling the input frequencies, we introduce a scaling factor \( \gamma > 1 \) such that:
\begin{equation}
\mathbf{z}^{(0)} = \sin\left( \gamma \Omega \mathbf{r} \right).
\end{equation}
Applying the same analysis, the output of the second layer becomes:
\begin{align}
z_m^{(1)} &= \operatorname{Im}\left\{ \exp\left( j \sum_{t=0}^{T-1} W_{m,t}^{(1)} \sin\left( \gamma \boldsymbol{\omega}_t^\top \mathbf{r} \right) \right) \right\} \nonumber \\
&= \operatorname{Im}\left\{ \prod_{t=0}^{T-1} \exp\left( j W_{m,t}^{(1)} \sin\left( \gamma \boldsymbol{\omega}_t^\top \mathbf{r} \right) \right) \right\} \nonumber \\
&= \operatorname{Im}\left\{ \prod_{t=0}^{T-1} \left( \sum_{s_t=-\infty}^{\infty} J_{s_t}\left( W_{m,t}^{(1)} \right) e^{j s_t \gamma \boldsymbol{\omega}_t^\top \mathbf{r}} \right) \right\} \nonumber \\
&= \sum_{\mathbf{s} \in \mathbb{Z}^T} \left( \prod_{t=0}^{T-1} J_{s_t}\left( W_{m,t}^{(1)} \right) \right) \sin\left( \gamma \sum_{t=0}^{T-1} s_t \boldsymbol{\omega}_t^\top \mathbf{r} \right).
\end{align}
By introducing the scaling factor \( \gamma \), we effectively scale the frequencies of the sinusoidal components in the output by \( \gamma \). When \( \gamma \) increases, the frequencies \( \gamma \sum_{t=0}^{T-1} s_t \boldsymbol{\omega}_t \) increase proportionally, extending the network's capacity to represent higher-frequency components. This adjustment enhances the network's ability to represent high-frequency information by broadening the range of frequencies it can model.

\subsection{ReLIFT and Residual Connections}
\begin{figure}[!thb]
    \centering
    \includegraphics[width=\linewidth]{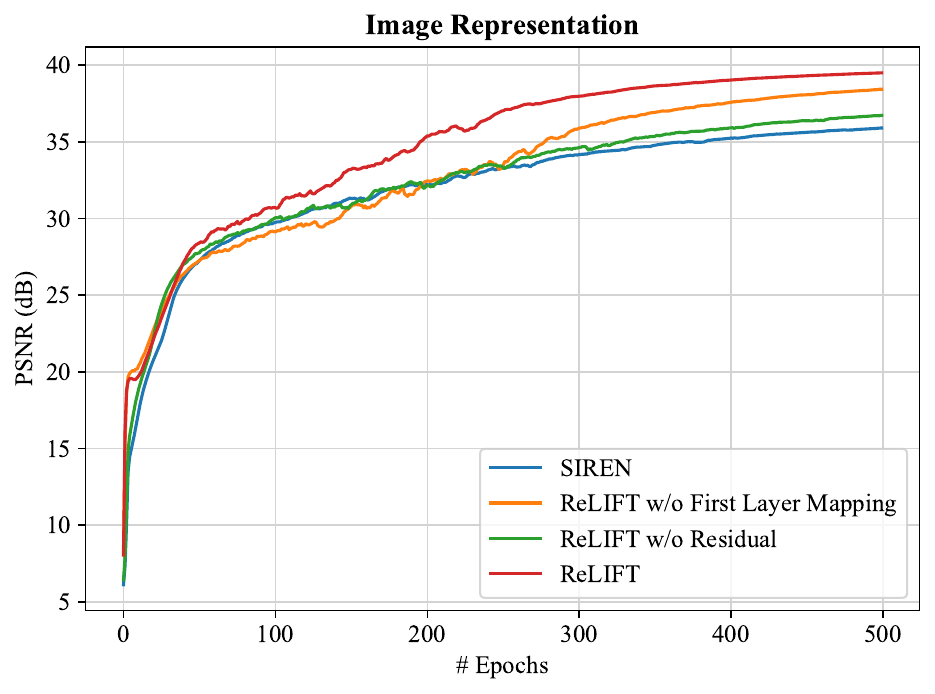}\\
    \caption{Comparison of convergence rates between the standard SIREN and ReLIFT on an image representation task. ReLIFT exhibits faster convergence and higher capacity. $\gamma$ is set to 2.}
    \label{fig:convergence_comparison}
\end{figure}

SIRENs have shown impressive abilities in representing complex signals using sinusoidal activations~\cite{sitzmann2020implicit}. However, they face a capacity-convergence gap when modeling high-frequency components. Although scaling the first layer \( \omega_0 \) can improve the network’s capacity for higher frequencies, it may not fully bridge this gap (see~\autoref{fig:convergence_comparison}). To overcome these limitations, we propose adding residual connections to our architecture, aiming to enhance both expressiveness and convergence.

\noindent In our proposed architecture, ReLIFT, we integrate residual connections into SIREN, where each layer adds its input to its output. This approach facilitates gradient flow, preserves low-frequency representations, and enables better modeling of higher frequencies. The network layers are defined as follows:
{\small
\begin{equation}
\begin{aligned}
    &\mathbf{z}^{(0)} = \sin\left(\gamma\omega_0 (W^{(0)} \mathbf{r}+ \mathbf{b}^{(0)})\right) \\
    &\mathbf{z}^{(l)} = \sin\left(\omega_0 (W^{(l)} \mathbf{z}^{(l-1)} + \mathbf{b}^{(l)})\right) \textcolor{blue}{+\mathbf{z}^{(l-1)}}, \; l = 1, \dots, L-2, \\
    &f(\mathbf{r}; \theta) = W^{(L)} \mathbf{z}^{(L-1)} + \mathbf{b}^{(L)}.
\end{aligned}
\end{equation}
}
Let's consider the first layer outputs:
\begin{equation}
\mathbf{z}^{(0)} = \sin\left( \gamma \Omega \mathbf{r} \right).
\end{equation}
The second layer with a residual connection is given by:
\begin{equation}
\mathbf{z}^{(1)} =  \sin\left( \mathbf{W}^{(1)} \sin\left( \gamma \Omega \mathbf{r} \right) \right) + \mathbf{z}^{(0)}.
\label{eq:second_layer_output}
\end{equation}
Considering the frequency component analysis using Bessel function identities (see \autoref{sec:revisiting}), the output of the second layer becomes:
\begin{align}
\label{eq:second_layer_final}
z_m^{(1)} &= \sum_{\mathbf{s} \in \mathbb{Z}^T} \left( \prod_{t=0}^{T-1} J_{s_t}\left( W_{m,t}^{(1)} \right) \right) \sin\left( \gamma \sum_{t=0}^{T-1} s_t \boldsymbol{\omega}_t^\top \mathbf{r} \right) \nonumber \\
 &\quad + \sin( \gamma \omega_{m}^{T} r).
\end{align}
Intuitively, the residual connection in our network plays a crucial role in ensuring that lower-order harmonics (fundamental frequencies) are robustly represented. Although the Bessel expansion contributes to higher-order harmonics, the residual connection allows the network to avoid depending solely on the Bessel functions to capture lower frequencies. By preserving fundamental frequencies through the residual connection, the network can adjust the weights \( W^{(1)} \) to enhance higher-order harmonics without risking the attenuation of fundamental components. Furthermore, residual connections provide shortcut paths for gradients, which significantly improves gradient flow during backpropagation~\cite{he2016deep}, mitigating issues related to vanishing or exploding gradients.

\begin{figure*}[ht]
    \centering
    \begin{multicols}{2}
        \begin{minipage}{\linewidth}
            \centering
            \includegraphics[width=0.9\linewidth]{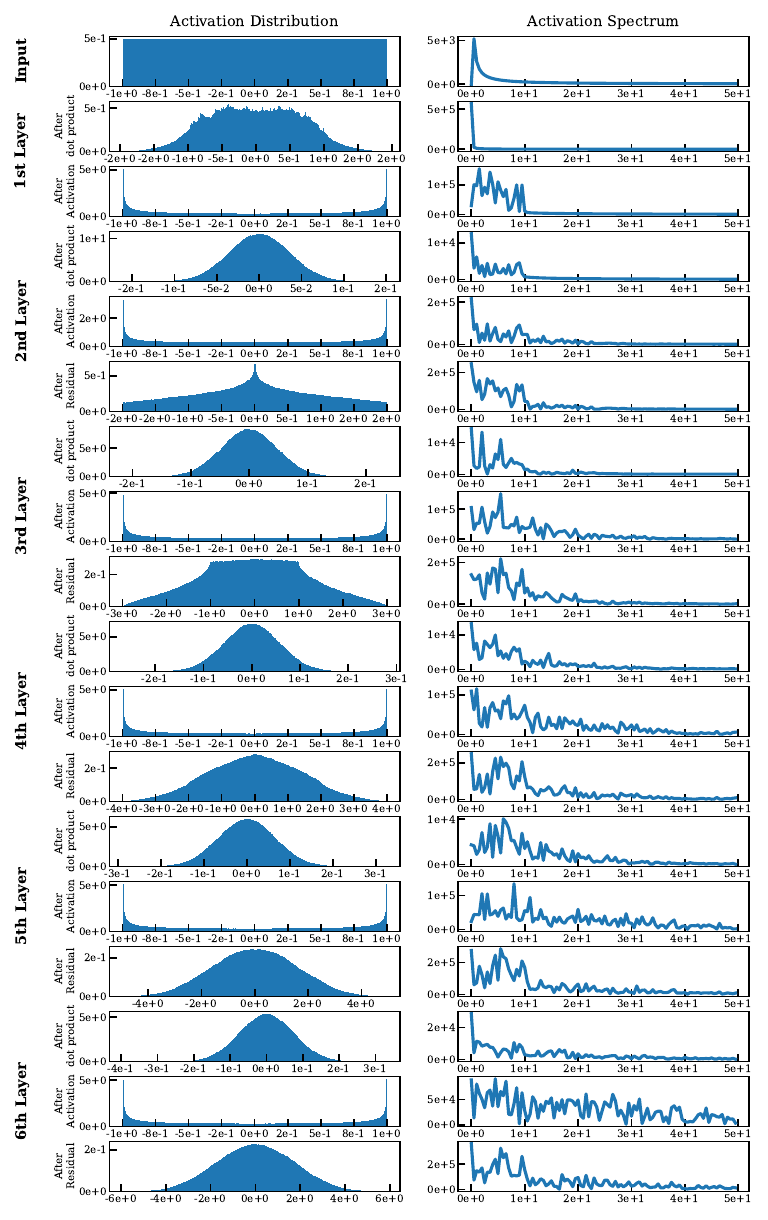}
            \par{(a) ReLIFT}
        \end{minipage}

        \begin{minipage}{\linewidth}
            \centering
            \includegraphics[width=0.98\linewidth]{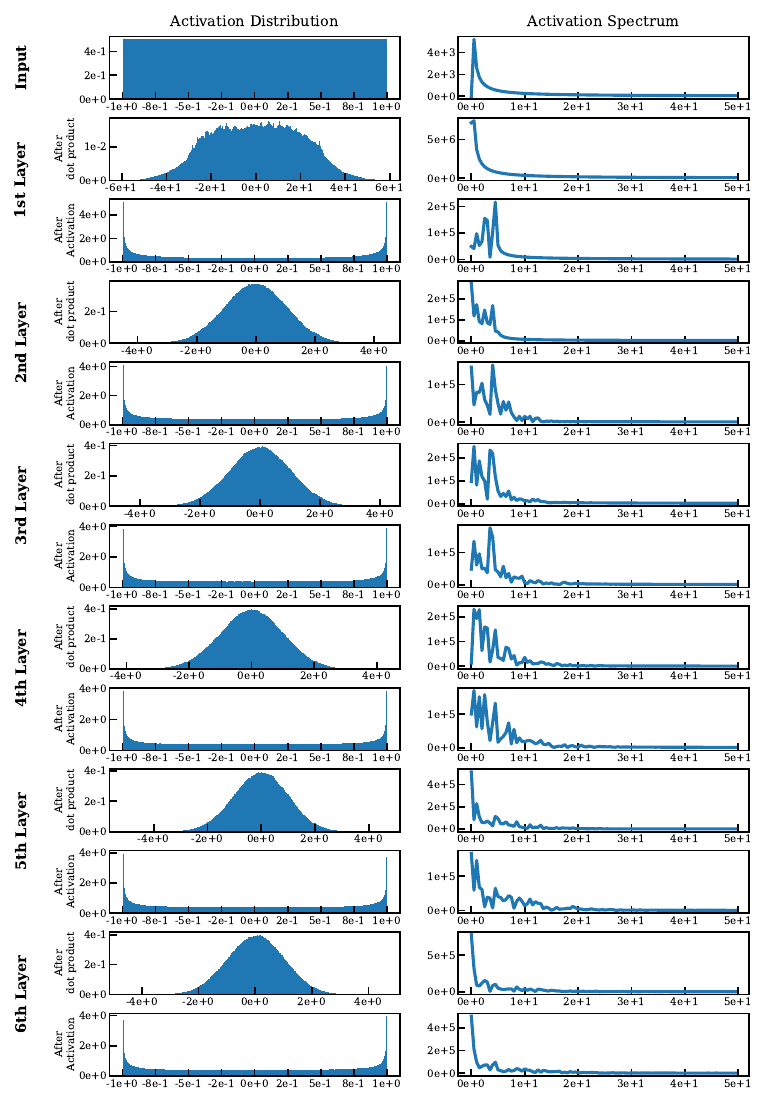}
            \par{(b) SIREN}
            \label{fig:activations_siren}
        \end{minipage}
    \end{multicols}
    \caption{Activation statistics comparison between ReLIFT and SIREN.}
    \label{fig:activation_comparison}
\end{figure*}

\noindent We conducted experiments comparing the standard SIREN model and our ReLIFT model, including ablations, on a single-image representation task with similar model configurations (a 5-layer MLP with a width of 256). As illustrated in~\autoref{fig:convergence_comparison}, ReLIFT achieved faster convergence and higher PSNR, demonstrating its effectiveness in reducing the gap in convergence and capacity. Additionally, we extend our method to various tasks, which are discussed in the following sections. We also present the activation statistics for ReLIFT and SIREN in~\autoref{fig:activation_comparison}. As shown, the maximum frequency increases with additional layers, allowing our model to capture a broader range of frequencies effectively.

\section{ReLIFT in the LIFT Framework}

We integrate ReLIFT into our LIFT framework, modifying the modulation approach of SIREN-based activations (as detailed in Section 3.2). Specifically, we introduce a residual connection to the output of the activation function and apply a scaling factor \( \gamma \) to the input layer frequency \( \omega_0 \).  To evaluate the impact of this approach, we perform a reconstruction task on the CelebA-HQ \( 64^2 \) dataset. Both the baseline and ReLIFT networks are trained using the configurations specified in \autoref{tab:configs}, highlighted in \colorbox{pink}{pink}. We also set $\gamma=2$, and trained both models for 120K iterations.

\noindent Our experiments reveal a marked improvement in convergence rate and reconstruction quality during the early stages of training with the ReLIFT model compared to the baseline LIFT configuration. Notably, at just 10K iterations, ReLIFT achieves a PSNR that is 2.3 points higher than LIFT, and by 20K iterations, ReLIFT reaches a PSNR of 38.21, outperforming all baselines, which typically requires 200K iterations to approach comparable results. Regarding computational efficiency, ReLIFT reaches 10K iterations in a wall time of 68 minutes and 20K in approximately 134 minutes, with a per-device batch size of 128 on 4 A40 GPUs. This efficiency demonstrates that, within only 134 minutes, ReLIFT can achieve high-fidelity reconstructions that surpass all current SOTA methods, making it a compelling model for rapid, high-quality image reconstruction.
\begin{figure}[t]
    \centering
    \includegraphics[width=\linewidth]{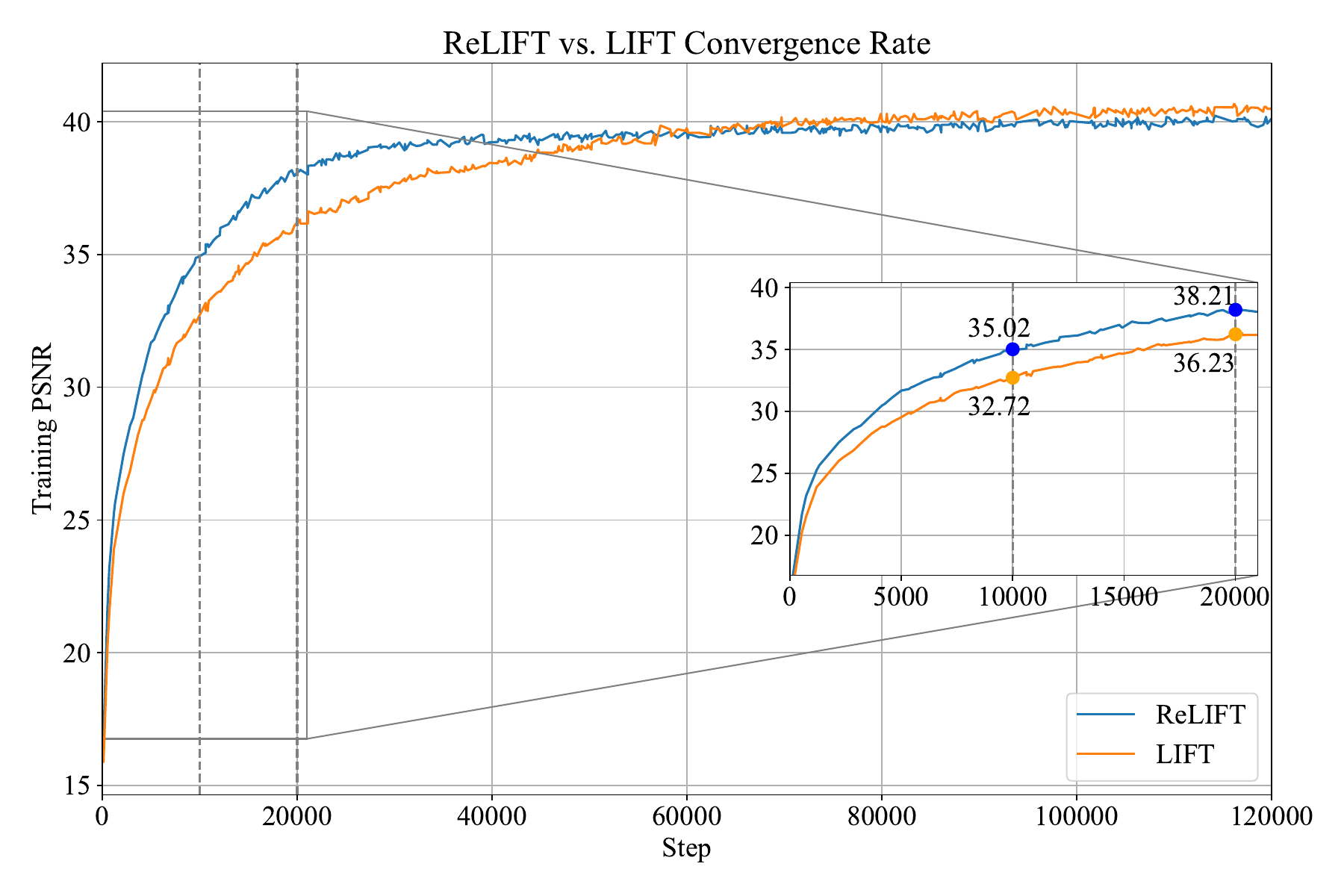}\\
    \vspace{-1em}
    \caption{Convergence rate comparison of ReLIFT and LIFT.}
\end{figure}

\section{Single Data Task Analysis}

To evaluate the effectiveness of our straightforward approach, ReLIFT, we extend our experiments to single data tasks, including \textbf{signal representation} and \textbf{inverse problems}. Our results demonstrate that ReLIFT can significantly reduce the convergence-capacity gap and achieve SOTA performance across all tasks, without introducing additional learnable parameters or requiring new activation functions. These findings highlight ReLIFT’s ability to surpass existing SOTA methods with a simple yet powerful adjustment, making it highly applicable across a range of tasks. 

\noindent We evaluate \texttt{ReLIFT} against six SOTA INRs: ReLU with positional encoding (\textbf{ReLU+P.E})~\cite{tancik2020fourier}, Fourier feature embedding (\textbf{FFN})~\cite{tancik2020fourier}, \textbf{SIREN}~\cite{sitzmann2020implicit}, Gaussian-based activation functions (\textbf{Gauss})~\cite{ramasinghe2022beyond}, wavelet activation functions (\textbf{WIRE})~\cite{saragadam2023wire}, \textbf{FINER}~\cite{liu2024finer}. To ensure a fair comparison, each INR is configured with the same network architecture, consisting of 3 hidden layers with 256 neurons per layer, and is trained using the Adam optimizer~\cite{kingma2014adam} and an L2 loss function between the network output and the ground truth. Other hyperparameters follow the specifications provided in the authors' open-source code and WIRE~\cite{saragadam2023wire}. Experiments run for 500 epochs, except for audio (1000 epochs) and occupancy (200 epochs). We also set $\gamma=2$ for ReLIFT.

\subsection{Signal Representations}

\subsubsection{Image}
\begin{figure*}[!h]
    \centering
    \includegraphics[width=0.86\textwidth]{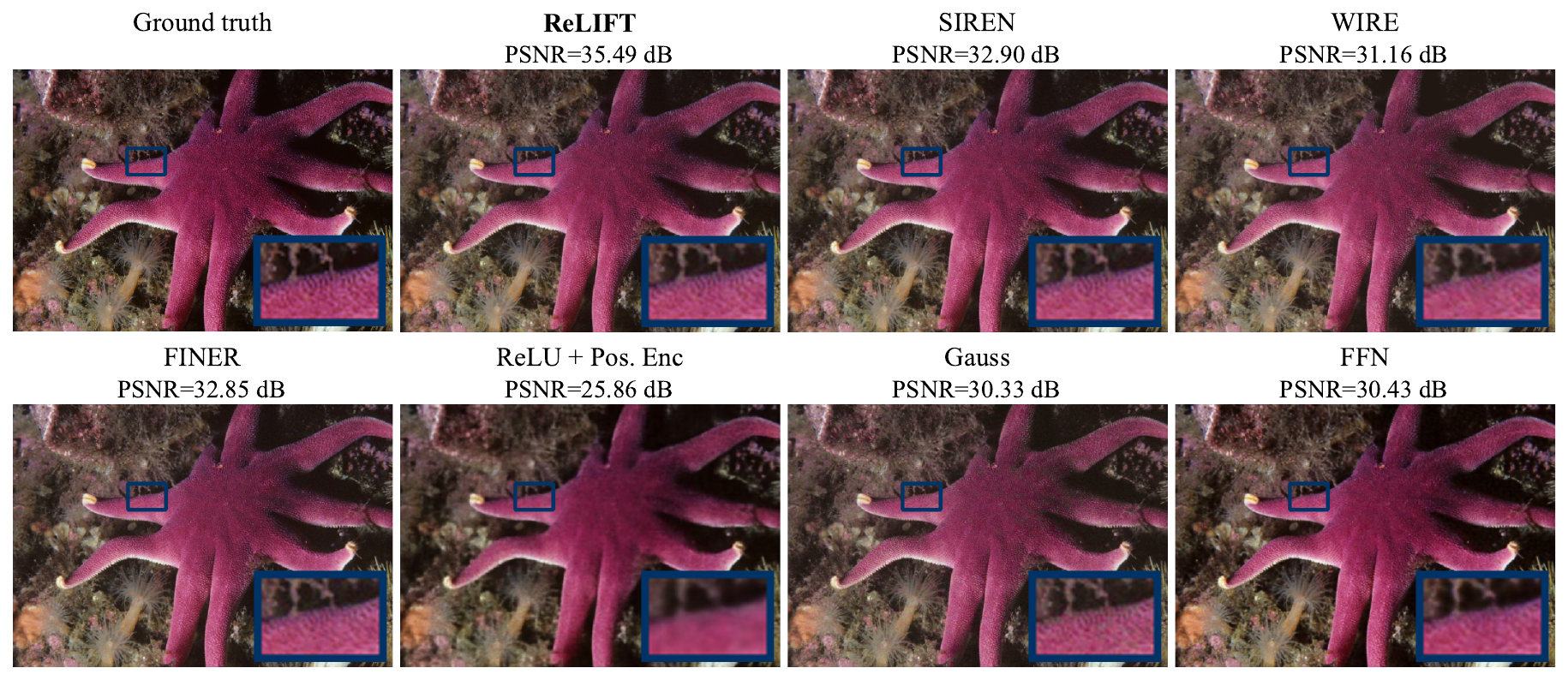}\\
    \includegraphics[width=0.86\textwidth]{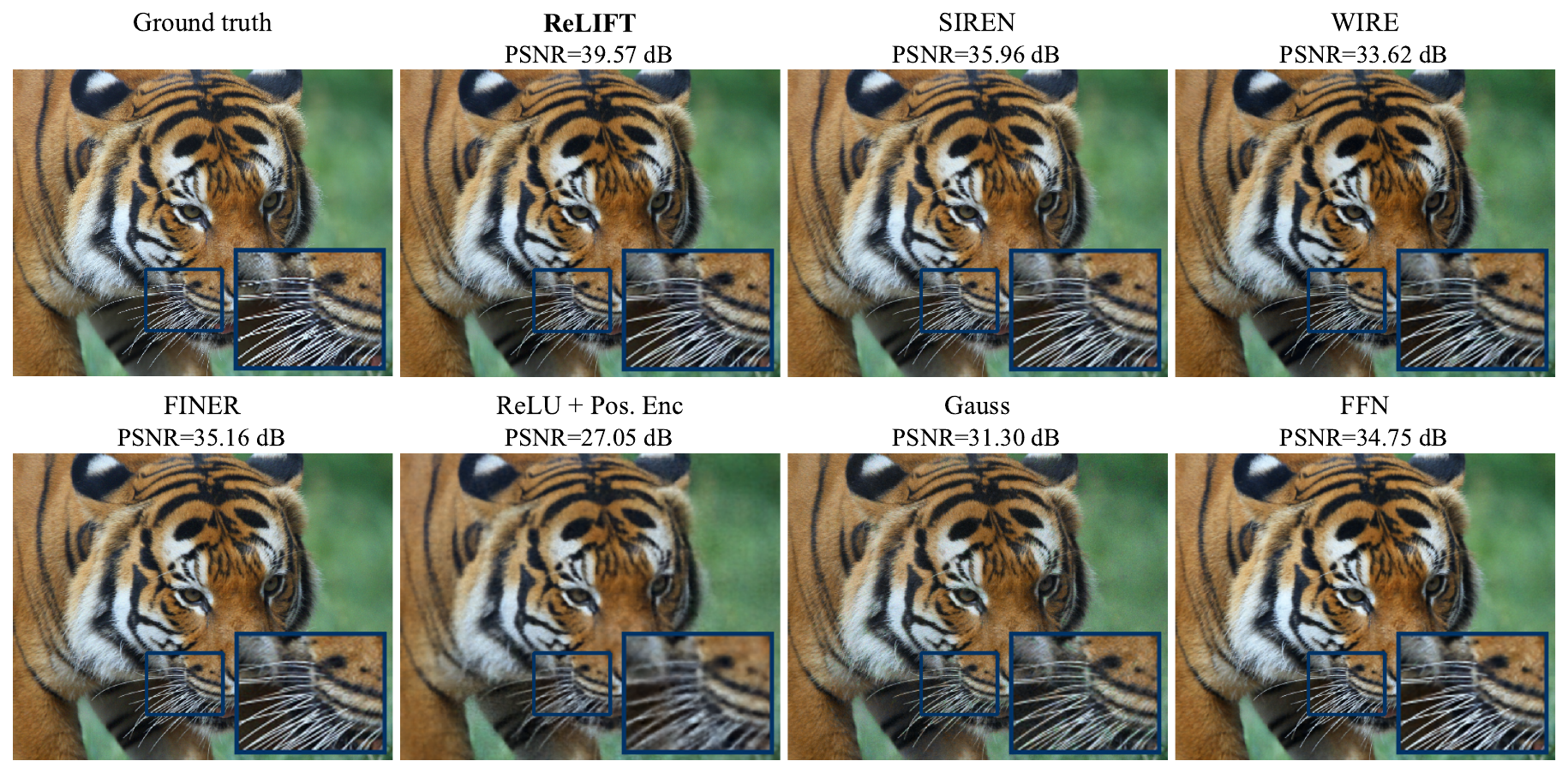}\\
    \includegraphics[width=0.86\textwidth]{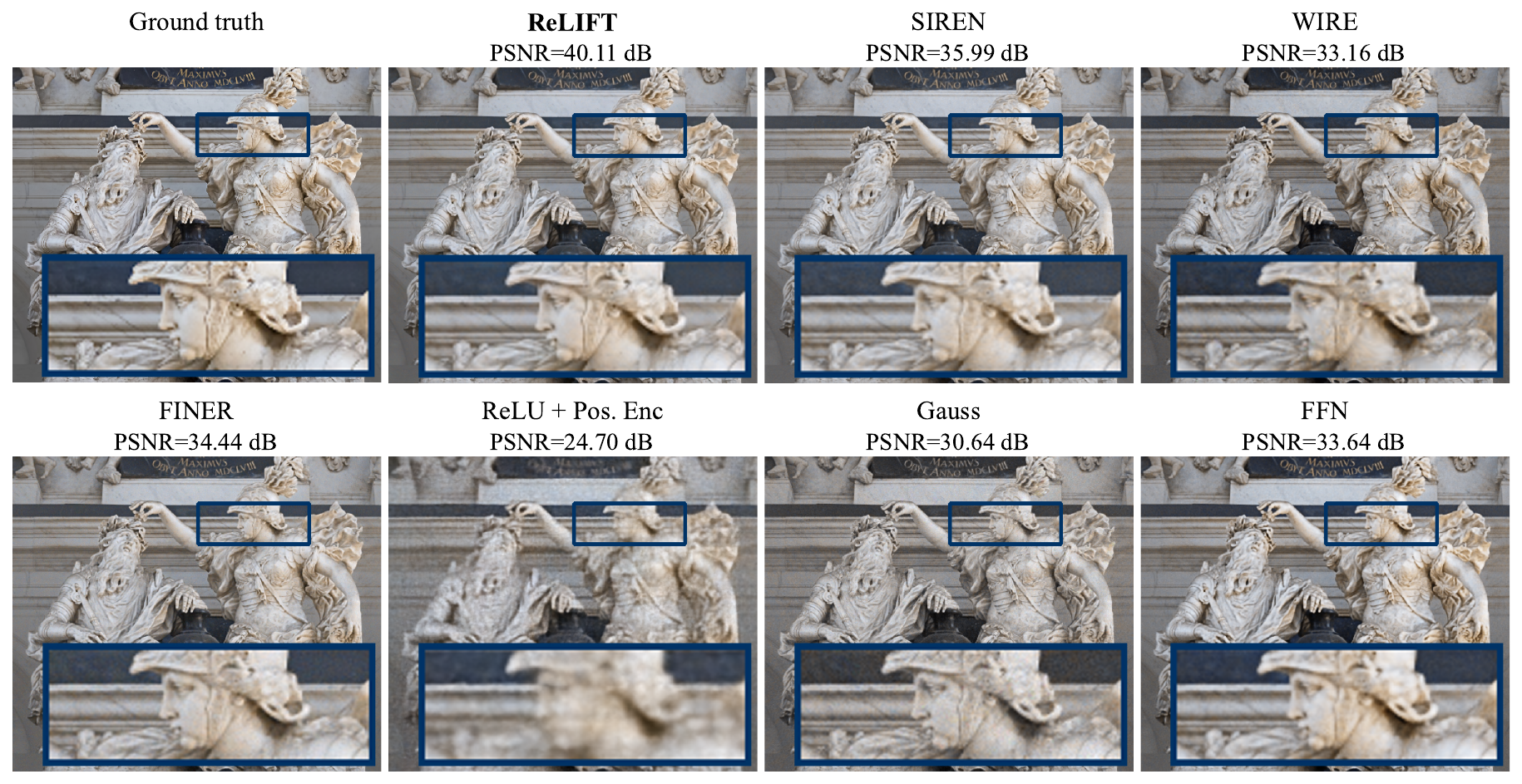}
    \caption{\textbf{Image representation:} PSNR comparisons of ReLIFT with SOTA models.}
    \label{fig:image_rep}
\end{figure*}

\textbf{Data.} For the image representation task, we use the DIV2K dataset~\cite{div2k}, with images downscaled to 1/4 of their original size. In~\autoref{fig:image_rep}, the first image (octopus) is resized from \(1404 \times 2040 \times 3\) to \(351 \times 510 \times 3\). The second image is trained at a resolution of \(411 \times 510 \times 3\), and the third at \(435 \times 510 \times 3\). 

\noindent\textbf{Analysis.} Given a 2D point \( (x, y) \), the INR learns a mapping function \( f: \mathbb{R}^2 \rightarrow \mathbb{R}^3 \) that outputs the RGB values. The results in~\autoref{fig:image_rep} highlight that ReLIFT consistently surpasses other INR methods in PSNR across various images, showcasing its superior reconstruction capability. For the first image (octopus), ReLIFT delivers sharper reconstructions, particularly in the highlighted area. In contrast, competing methods such as ReLU+P.E., FFN, and WIRE produce noticeably blurrier outputs, and Gauss introduces color artifacts. ReLIFT improves the PSNR by 2.59 and 2.64 over SIREN and FINER, the second and third-best methods, respectively. Similarly, for the second image (tiger), ReLIFT achieves a PSNR increase of 3.61 and 4.41 over SIREN and FINER, demonstrating the method's reliability across different images. In the final, higher-resolution image, ReLIFT achieves a PSNR of 40.11, surpassing its closest competitor by a margin of 4.12, further reinforcing its capability for high-fidelity reconstructions.

\subsubsection{Occupancy Volume}
\textbf{Data.} We evaluate our approach using 4 shapes from a public dataset~\cite{stanford3d,muller2022instant}. For each shape, we create occupancy volumes by point sampling on a \(512 \times 512 \times 512\) grid, assigning a value of 1 to voxels within the shape and 0 to those outside. 

\begin{figure*}[!h]
    \centering
    \includegraphics[width=\textwidth]{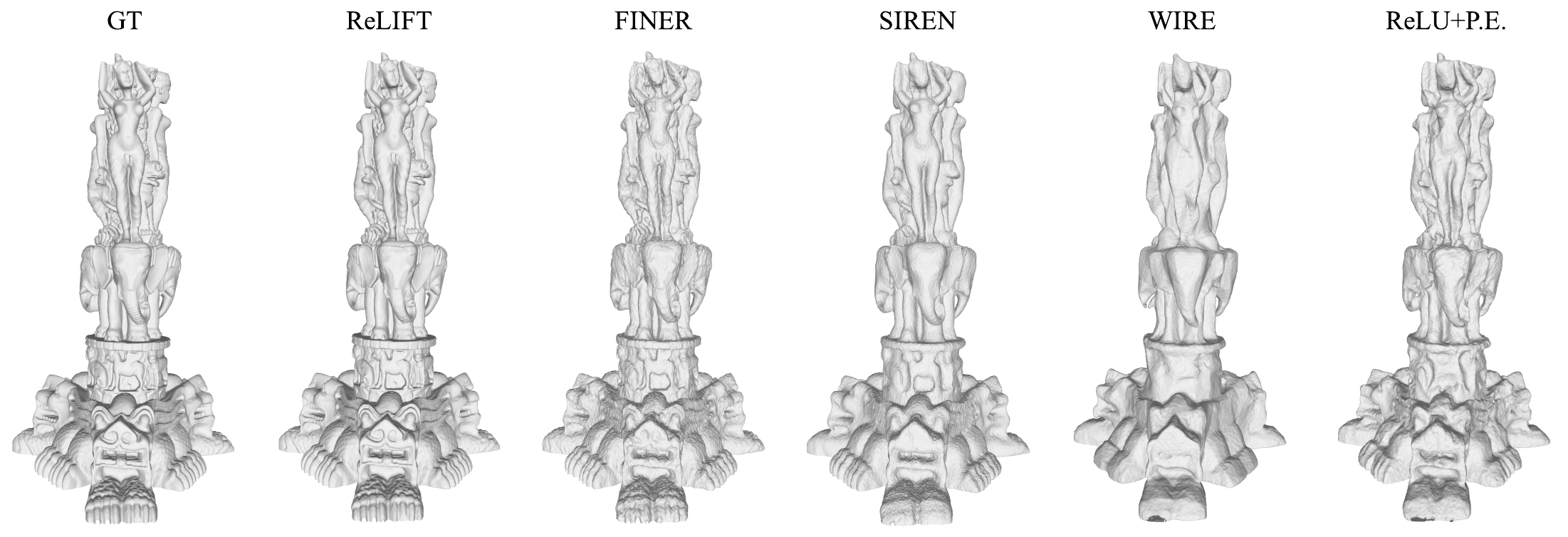}\\
    \caption{\textbf{Shape representation:} Qualitative comparisons of ReLIFT with SOTA models. }
    \label{fig:sdf_rep}
\end{figure*}

\noindent\textbf{Analysis.} Given a 3D point \( (x, y, z) \), the INR learns a mapping function \( f: \mathbb{R}^3 \rightarrow \mathbb{R} \) that outputs the signed distance field (SDF) values. The quantitative comparisons in~\autoref{tab:signed_distance_field} demonstrate the effectiveness of our approach, ReLIFT, in representing SDF across various shapes. In terms of Intersection over Union (IOU), ReLIFT consistently outperforms other methods on all tested shapes: Armadillo, Dragon, Lucy, and Thai Statue. On average, ReLIFT achieves the highest IOU (0.9963), surpassing the second-best method, FINER, with an average of 0.9944. The qualitative comparison is shown in~\autoref{fig:sdf_rep} for the Thai statue. ReLIFT achieves the best overall performance, capturing both high-frequency details and smooth transitions, closely aligning with the ground truth. FINER retains broader structural features but sacrifices finer details, leading to coarser outputs. SIREN performs well in smooth regions but introduces artifacts and struggles with intricate features, resulting in less accurate reconstructions. WIRE produces overly smoothed approximations with significant detail loss, making it suitable only for coarse representations. ReLU+P.E. benefits from positional encoding to improve spatial structure over WIRE and SIREN but does not reach the fidelity and precision of ReLIFT and FINER.

\begin{table}[h]
    \centering
    \caption{Quantitative comparisons of SDF representations. Each cell is color-coded to represent the \colorbox{red}{\kern-\fboxsep best\kern-\fboxsep}, \colorbox{orange}{\kern-\fboxsep second-best\kern-\fboxsep}, and \colorbox{yellow}{\kern-\fboxsep third-best\kern-\fboxsep} performance.}
    \resizebox{\linewidth}{!}{%
    \begin{tabular}{ccccccc}
        \toprule
         & Methods & Armadillo & Dragon & Lucy & Thai Statue & Avg. \\ \midrule
        \multirow4{*}{\rotatebox{90}{\textbf{IOU $\uparrow$}}} & ReLU+P.E. & \cellcolor{yellow}0.9966 & {\cellcolor{yellow}}0.9963 & {\cellcolor{yellow}}0.9919 & 0.9906 & {\cellcolor{yellow}}0.9939 \\ 
        & SIREN & 0.9968\cellcolor{orange} & {\cellcolor{orange}}0.9969 & 0.9881 & \cellcolor{orange}0.9934 & 0.9938 \\ 
        & WIRE & 0.9677 & 0.9724 & 0.9705 & 0.9484 & 0.9648 \\ 
        & FINER & 0.9958 & 0.9945 & {\cellcolor{orange}}0.9955 & \cellcolor{yellow}0.9919 & {\cellcolor{orange}}0.9944  \\ 
        \midrule
        & \textbf{ReLIFT} & {\cellcolor{red}}0.9974 & 0.9975{\cellcolor{red}} & 0.9960{\cellcolor{red}} & {\cellcolor{red}}0.9943 & 0.9963\cellcolor{red} \\ \bottomrule
    \end{tabular}%
    }
    \label{tab:signed_distance_field}
\end{table}

\subsubsection{Audio Representations}
\textbf{Data.} For our audio representation task, we use the initial 7 seconds of Bach’s Cello Suite No. 1: Prelude~\cite{sitzmann2020implicit}, sampled at a rate of 44100 Hz.

\begin{figure*}[!th]
    \centering
    \includegraphics[width=\textwidth]{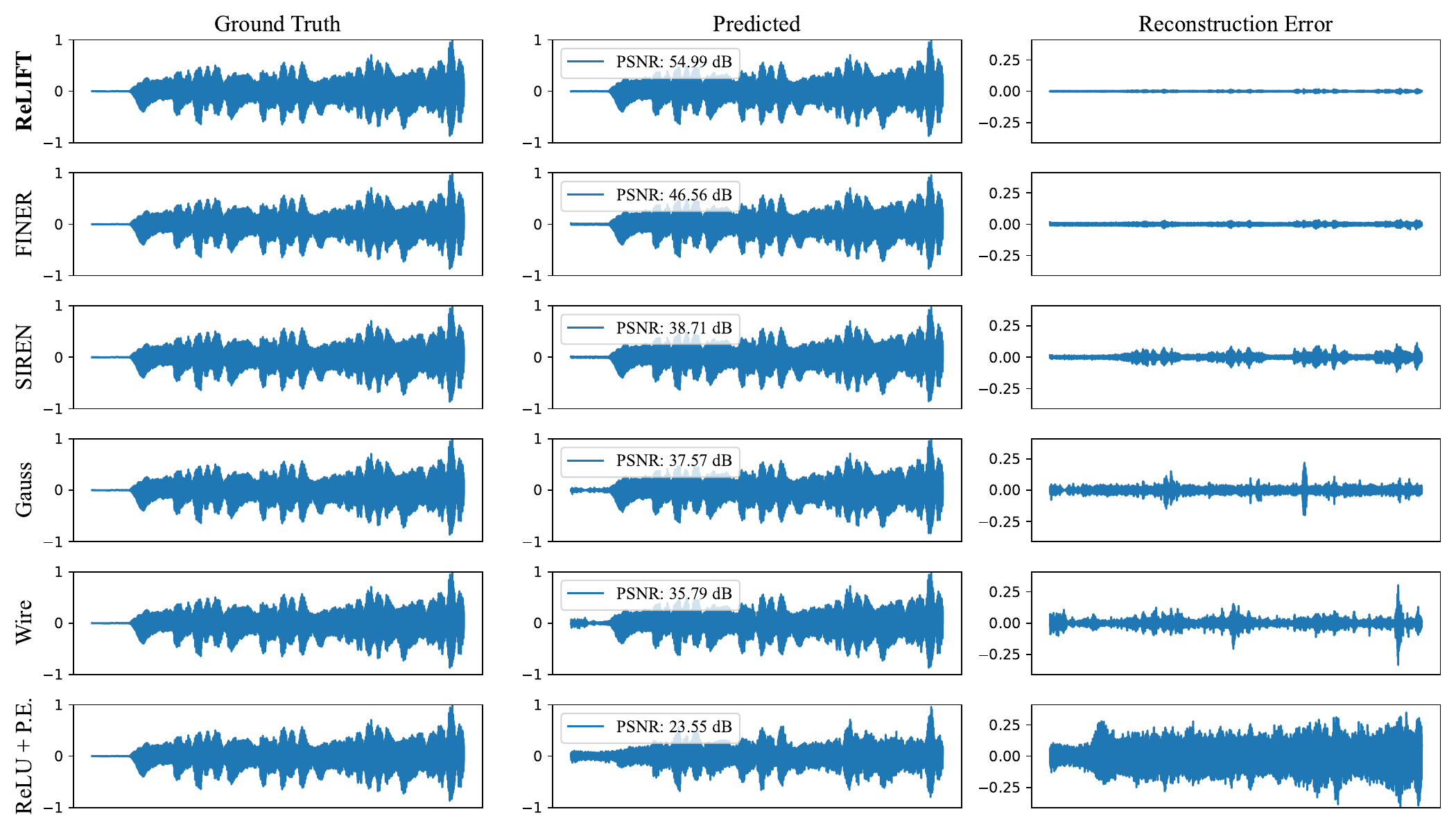}\\
    \caption{\textbf{Audio representation:} PSNR and reconstruction error comparisons of ReLIFT with SOTA models.}
    \label{fig:audio_rep}
\end{figure*}

\noindent\textbf{Analysis.} We evaluate ReLIFT's performance against other methods to assess its effectiveness in audio signal representation. Given a 1D point, the INR learns a mapping function \( f: \mathbb{R} \rightarrow \mathbb{R} \). ReLIFT leverages its residual capabilities and scaling factor, allowing us to use a higher frequency scaling value \( w_0 \). For ReLIFT, we set the first layer \( \omega_0 \) to 10000, scaled by \( \gamma = 2 \), and use hidden layers with \( \omega_0 = 90 \). In contrast, the original SIREN architecture suffers when high values of \( \omega_0 \) are used. According to Neural Tangent Kernel (NTK) analysis~\cite{jacot2018neural} in~\cite{yuce2022structured}, excessively large \( \omega_0 \) values can lead to a poorly conditioned NTK, where certain eigenvalues become too small. This issue hinders effective learning, resulting in SIREN’s poor performance on high-frequency representations. By addressing this challenge, ReLIFT achieves a PSNR of 54.99 dB, outperforming SIREN and other methods, with FINER as the next best at 46.56 dB—a difference of +8.43 dB (see~\autoref{fig:audio_rep}). The periodic nature of audio signals allows ReLIFT to produce a clear and accurate representation, similar to SIREN, but without the background noise issues. ReLIFT quickly reaches a low-error representation, while methods like Gauss, WIRE, and ReLU+P.E. introduce more noticeable distortion during playback. SIREN and FINER reduce this problem up to a point, but background noise is still present. Overall, ReLIFT performs best in minimizing error, as reflected in its PSNR value and reconstruction error.

\subsection{Inverse Problems}

\subsubsection{Image Super-resolution}
\textbf{Data.} An image from the DIV2K dataset~\cite{div2k} is used, with downsampling applied from an original resolution of 1356 × 2040 × 3 by scaling factors of 1/2, 1/4, and 1/6.

\begin{figure*}[!thb]
    \centering
    \includegraphics[width=\textwidth]{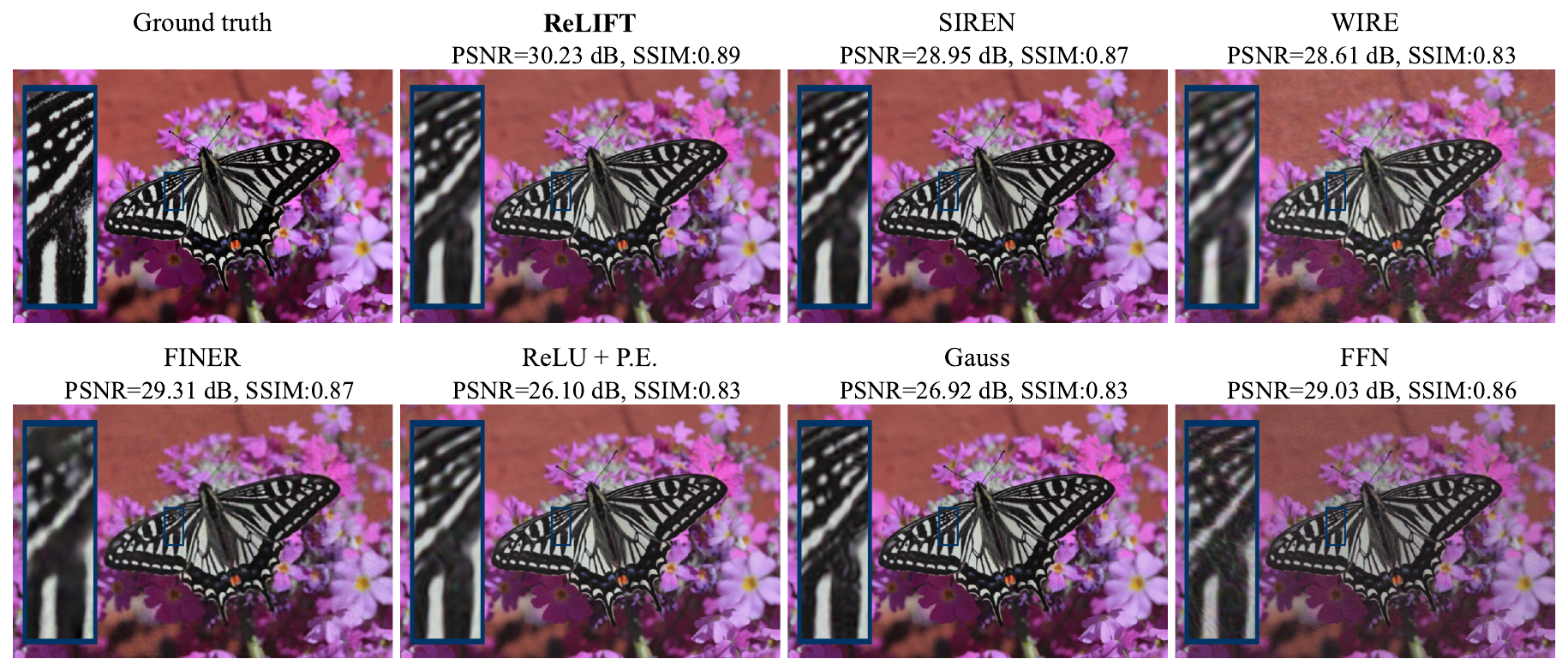}\\
    \caption{\textbf{Image Super-resolution:} PSNR and SSIM comparisons of a $4\times$ single image super-resolution between ReLIFT and SOTA models.}
    \label{fig:super_res}
\end{figure*}

\noindent\textbf{In super-resolution.} INRs serve as effective interpolants, using their natural strengths and inherent biases to enhance performance. To test this idea, we performed \(1\times\), \(2\times\), \(4\times\), and \(6\times\) super-resolution experiments on an image. The results in \autoref{tab:exp_sr_results} show that ReLIFT achieves the highest PSNR and SSIM values at every super-resolution level, outperforming other top methods. For instance, ReLIFT reaches a PSNR of 34.30 and an SSIM of 0.94 at \(1\times\) resolution and keeps a strong performance up to \(6\times\) resolution with a PSNR of 27.28 and an SSIM of 0.85. Visual comparisons~\autoref{fig:super_res} also show that ReLIFT preserves sharper details, while other methods tend to produce blurrier results, highlighting ReLIFT's ability to maintain high-quality reconstruction.
\begin{table}[!tbh]
    \centering
    \vspace{-0.5em}
    \caption{ReLIFT vs. SOTAs in super-resolution.}
    \vspace{-0.5em}
    \resizebox{1\linewidth}{!}{
    \begin{tabular}{lcclcclcclcc} 
    \toprule
    \multirow{2}{*}{Methods} & \multicolumn{2}{c}{1$\times$} &  & \multicolumn{2}{c}{2$\times$} &  & \multicolumn{2}{c}{4$\times$} &  & \multicolumn{2}{c}{6$\times$} \\
     & PSNR & SSIM &  & PSNR~ ~ & SSIM &  & PSNR & SSIM &  & PSNR & SSIM \\ 
    \midrule
    Gauss & 30.32 & 0.79 &  & 29.15 & 0.86 &  & 26.92 & 0.83 &  & 25.17 & 0.81 \\
    FFN & {\cellcolor{yellow}} 32.83 &  0.90 {\cellcolor{orange}} &  & 29.28 & 0.85 &  & {\cellcolor{yellow}}29.03 & 0.86 &  & {\cellcolor{orange}}27.05 & {\cellcolor{orange}}0.84 \\
    ReLU P.E. & 32.46 & 0.87 &  & 30.41 & 0.88 &  & 26.10 & 0.83 &  & 24.61 & 0.81 \\
    WIRE & 31.63 & 0.85 &  & 31.28 & 0.86 &  & 28.61 & 0.83 &  & 24.76 & 0.70 \\
    SIREN & 32.02 & 0.87 &  & {\cellcolor{yellow}} 31.55 & {\cellcolor{yellow}} 0.89 &  & 28.95 & 0.87{\cellcolor{orange}} &  & 26.37 & {\cellcolor{orange}}0.84 \\
    FINER & 34.35{\cellcolor{red}} & 0.89 {\cellcolor{yellow}} &  &32.45{\cellcolor{orange}} & 0.90{\cellcolor{red}}  &  & 29.31 {\cellcolor{orange}} & 0.87{\cellcolor{orange}} &  & {\cellcolor{yellow}}26.94 & 0.83  \\
    \midrule
    \textbf{ReLIFT} & {34.30}{\cellcolor{orange}} & {0.94}{\cellcolor{red}} & & {33.07}{\cellcolor{red}} & {0.90}{\cellcolor{red}} & & {\cellcolor{red}} {30.23} & {\cellcolor{red}} {0.89} &  & {27.28}{\cellcolor{red}} & {0.85} {\cellcolor{red}} \\
    \bottomrule
    \end{tabular}
    }
    \vspace{-1.5em}
    \label{tab:exp_sr_results}
\end{table}

\subsubsection{Image Denoising}
\noindent\textbf{Data.} We use an image from the DIV2K dataset \cite{div2k}, downsampled by a factor of 1/4 from an original resolution of $1356 \times 2040 \times 3$ to $339 \times 510 \times 3$. To simulate realistic sensor noise, we apply photon and readout noise, where each pixel is affected by independent Poisson random variables. The mean photon count ($\tau$) is set to 40, and the readout count is fixed at 2.

\begin{figure*}[]
    \centering
    \includegraphics[width=\textwidth]{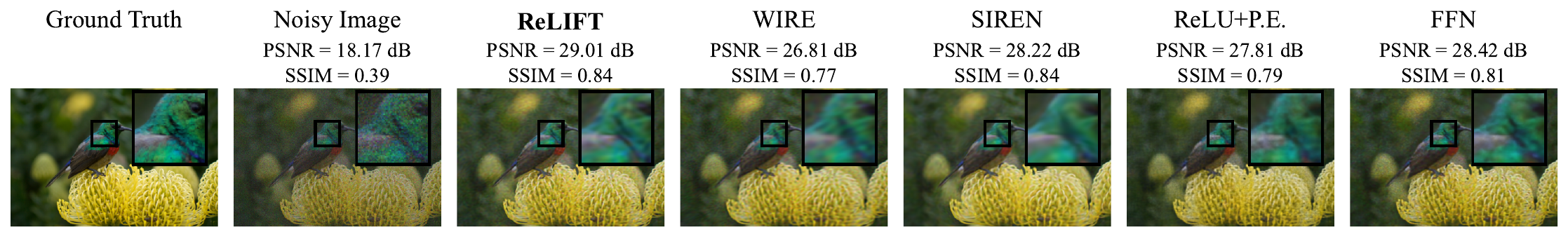}\\
    \caption{\textbf{Image Denoising:} PSNR and SSIM comparisons between ReLIFT and SOTA models.}
    \label{fig:denoising}
\end{figure*}

\noindent\textbf{Analysis.} We demonstrate ReLIFT’s effectiveness in tackling inverse problems, especially in image denoising, by leveraging its inductive bias and robustness to noise. To manage high-frequency noise patterns in noisy images, we set the first layer scaling parameter to $\omega_0 = 10$, which helps ReLIFT maintain a balance between low- and high-frequency information—a configuration we also apply to SIREN and WIRE for comparison. As shown in \autoref{fig:denoising}, ReLIFT achieves significant improvements, including a PSNR gain of +10.84 dB and an SSIM increase of 0.45 over the original noisy image. ReLIFT effectively preserves image details while reducing noise artifacts, as evident in the zoomed-in areas, where both SIREN and WIRE exhibit over-smoothed details. While ReLU-based networks primarily capture low-frequency information, ReLIFT excels at balancing low- and high-frequency features benefiting from its optimized scaling and residual connections.

\subsubsection{Inpainting}
\begin{figure*}[!thb]
    \centering
    \includegraphics[width=\textwidth]{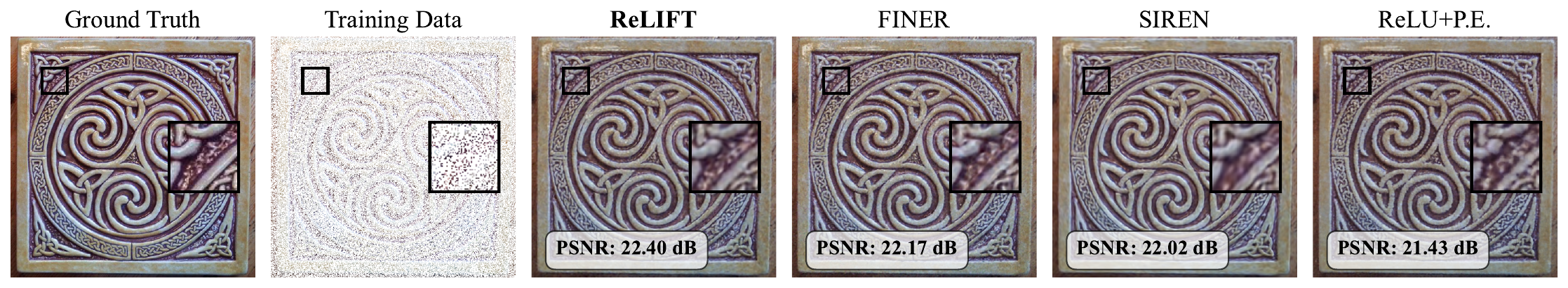}\\
    \caption{\textbf{Inpainting:} PSNR comparison between LIFT and SOTA models on 25\% of the pixels in a $572 \times 582 \times 3$ image.}
    \label{fig:inpainting}
\end{figure*}

\textbf{Data.} For the inpainting experiment, we use a \(572 \times 582 \times 3\) image, where 25\% of the pixels are masked as shown in the "Training Data" column in \autoref{fig:inpainting}. The masked pixels serve as missing data points that the models aim to reconstruct, while the remaining pixels provide context for the inpainting process. This setup tests each model’s ability to restore missing information while preserving image quality.

\noindent\textbf{Analysis.} In \autoref{fig:inpainting}, we compare the performance of ReLIFT with other SOTA methods, including FINER, SIREN, and ReLU+P.E., on the inpainting task. ReLIFT achieves the highest PSNR at 22.40 dB, outperforming FINER (22.17 dB), SIREN (22.02 dB), and ReLU+P.E. (21.43 dB). This improvement demonstrates ReLIFT’s superior capability to restore missing details with greater accuracy. In the zoomed-in regions, we observe that ReLIFT maintains sharper edges and textures compared to other methods, which tend to produce more blurred or smoothed reconstructions. FINER follows closely behind ReLIFT in terms of PSNR, showing reasonable inpainting quality but slightly softer details. SIREN and ReLU+P.E. further lag in performance, with more pronounced blurring in the reconstructed areas, suggesting less effective handling of fine textures and edges. ReLIFT's advantage stems from its design, which effectively balances high- and low-frequency features, allowing it to capture both broad structures and finer details in the inpainting process.


\section{ReLIFT Spectral Bias}
\begin{figure}[!thb]
    \centering
    \includegraphics[width=\linewidth]{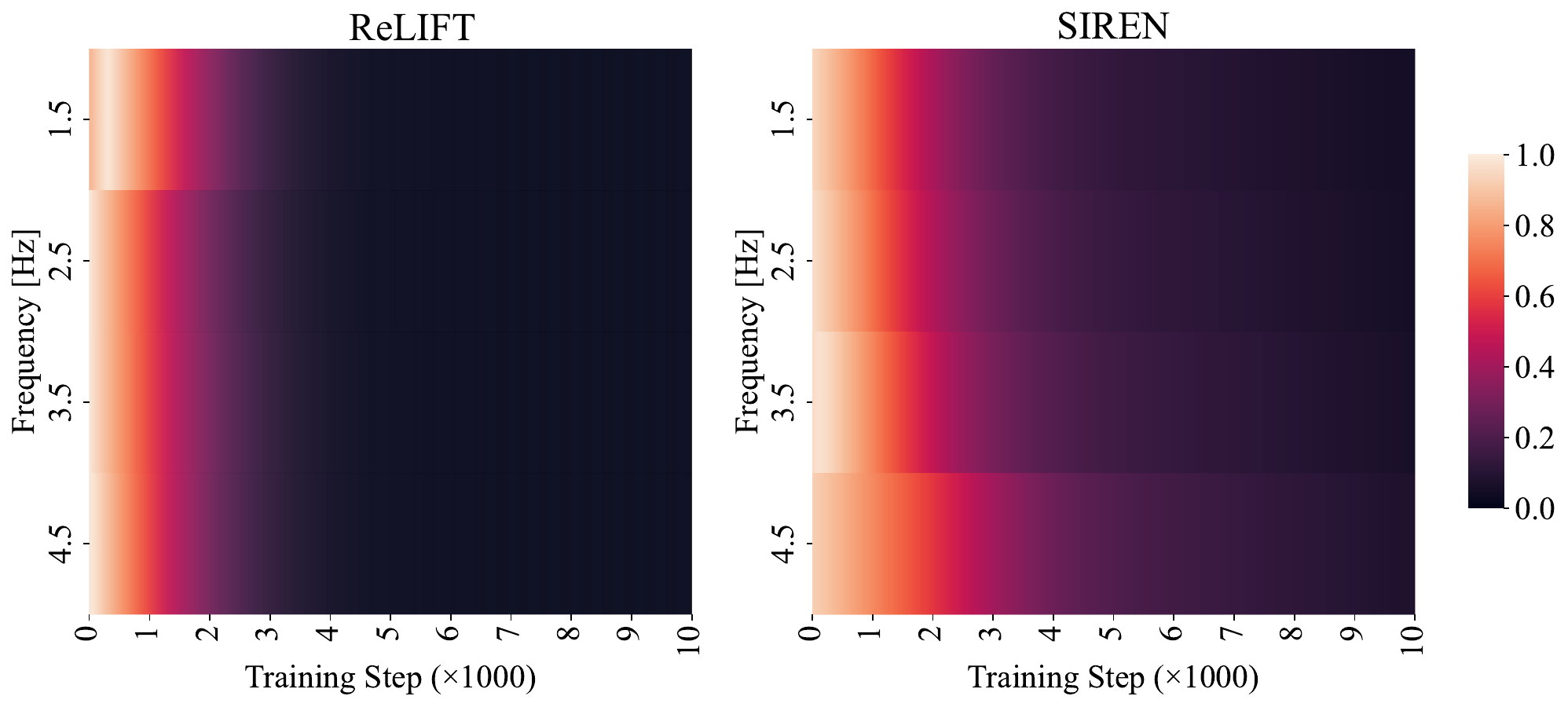}\\
    \caption{Frequency learning comparison between SIREN and ReLIFT. The x-axis shows training steps, the y-axis indicates frequency, and the color represents relative approximation error.}
    \label{fig:spectral_bias}
\end{figure}
Previous work by Rahaman et al.~\cite{rahaman2019spectral} has shown that MLP-based networks exhibit a spectral bias, where lower frequency components are learned more rapidly than higher ones. To investigate this spectral bias within our network, we adopted~\cite{shi2024improved} experimental framework using a 1D periodic function composed of four primary frequencies, as defined in~\autoref{eq:function}. The function \( f(x) \) was sampled at 300 points over the interval \([-1, 1]\).
{\small
\begin{equation}
    f(x) = 2R\left( \frac{\sin(3\pi x) + \sin(5\pi x) + \sin(7\pi x) + \sin(9\pi x)}{2} \right)
    \label{eq:function}
\end{equation}}
In this equation, \( R \) represents a rounding function that introduces discontinuities, thereby increasing the complexity of the training process. Our network architecture consists of a multilayer perceptron (MLP) with three hidden layers, each containing 128 neurons. We set the initial frequency parameter \( \omega_0 = 5 \) for both the SIREN model and our proposed method, ReLIFT, with a scaling factor \( \gamma = 2 \).

\noindent To evaluate the effectiveness of our approach, we trained both SIREN and ReLIFT on the defined function and compared their frequency learning dynamics. As illustrated in~\autoref{fig:spectral_bias}, SIREN demonstrates a spectral bias by quickly learning the low-frequency components while struggling to capture the high-frequency ones. In contrast, our ReLIFT method successfully balances the learning of both low and high-frequency components, thereby mitigating the spectral bias inherent in standard MLP-based networks.

\section{ReLIFT Layer Visualization}
\begin{figure*}[!thb]
    \centering
    \includegraphics[width=\textwidth]{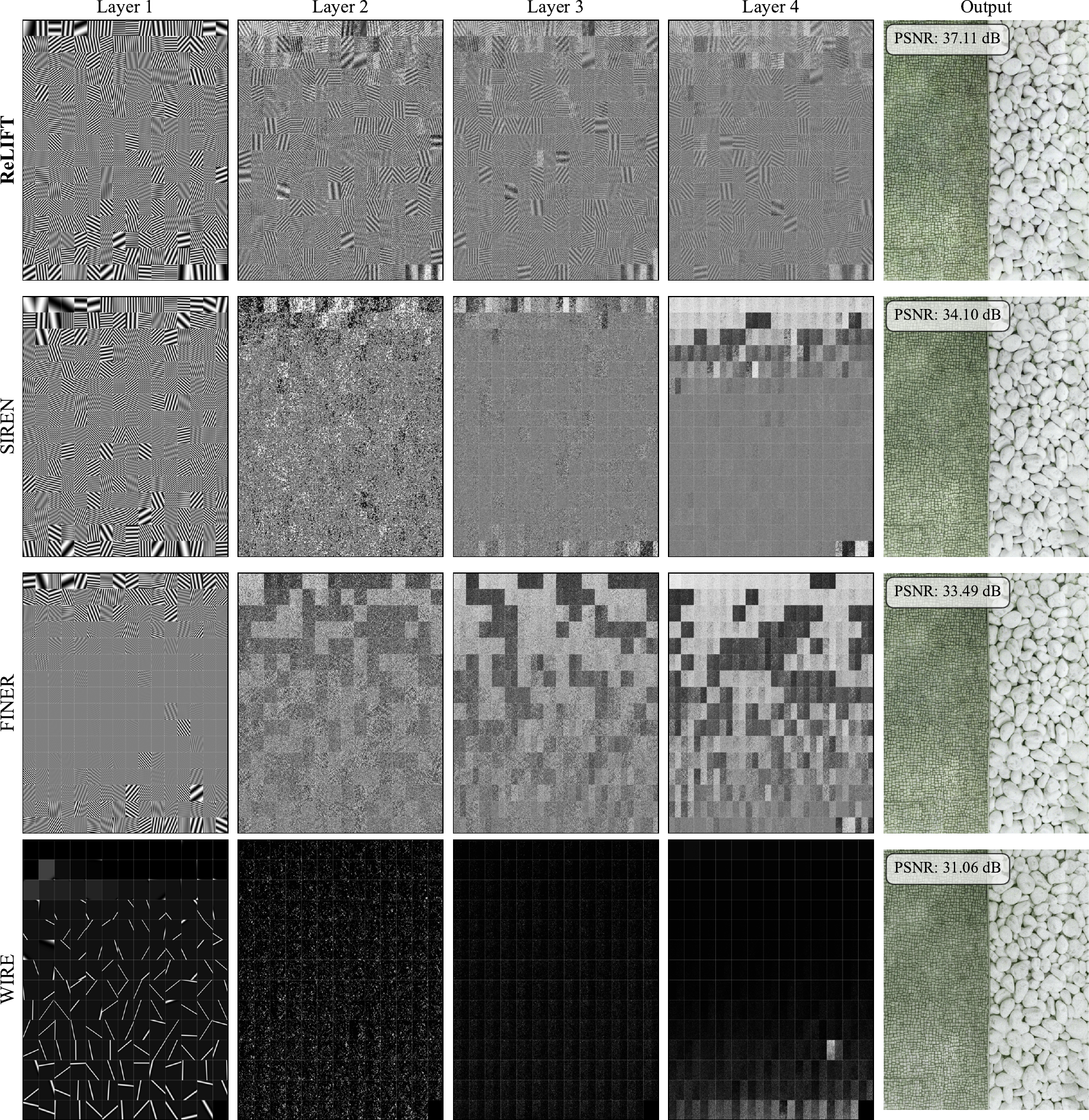}\\
    \caption{Visualization of hidden layer outputs.}
    \label{fig:layer_vis}
\end{figure*}

The layer visualization for ReLIFT, SIREN, FINER, and WIRE are depicted in~\autoref{fig:layer_vis}. We analyze each method below.
\subsection{ReLIFT}
ReLIFT displays structured and coherent spatial patterns across all layers, indicating strong feature retention. The patterns remain intricate even in the deeper layers, suggesting that ReLIFT's design allows it to capture and retain fine-grained details across the network depth. The PSNR of 37.11 dB aligns with the visual clarity of the patterns, as LIFT seems to retain more information with minimal degradation in detail. 

\subsection{SIREN:}
SIREN starts with sharp, detailed patterns in the initial layers, but these become progressively noisier and less defined in the later layers. While the initial layers retain significant variance, later layers lose high-frequency details. This could be due to the sinusoidal activation potentially introducing noise or leading to over-smooth representations in deeper layers. The PSNR of 34.10 dB reflects this slight degradation in detail. Although SIREN can capture complex features initially, it may struggle to maintain high fidelity across multiple layers, making it somewhat less effective than ReLIFT for fine-detail learning.

\subsection{FINER:}
FINER shows larger, blocky patterns that lose resolution as layers deepen, indicating a shift towards lower frequencies and a loss of spatial detail. This pattern suggests that FINER might prioritize coarse information, sacrificing fine-grained structure in favor of more generalized features. Due to the blocky representations, the filters in FINER likely have less variation in high frequencies, which is why the visualizations appear more uniform and less detailed compared to ReLIFT. With a PSNR of 33.49 dB, FINER generates an output that is less detailed and slightly blurry. 

\subsection{WIRE:}
WIRE quickly loses visible structure across layers, with activations fading into dark, nearly uniform representations by the later layers. This rapid loss of detail suggests that WIRE lacks mechanisms for retaining spatial structure across depth, perhaps due to an architecture that does not prioritize high-frequency detail retention. The lack of high-variance components leads to flatter, less informative patterns, which contribute to lower-quality visualizations and limited information retention. The PSNR of 31.06 dB reflects significant degradation in detail, resulting in a final output that is overly smooth and lacking in fidelity.

\section{Limitations and Discussion}
While our approach delivers promising results, certain limitations are worth discussing. In image-generation experiments, the generated outputs exhibit slight blurriness, which we hypothesize arises from the use of the sinusoidal activation function, potentially leading to over-smoothed representations. For 3D generation, we believe that fine-tuning the hyperparameters of the ADM model could further improve sample quality and evaluation scores. Compared to global-based methods that struggle with high-resolution generation, our hierarchical design effectively integrates global context with fine-grained local details. This multi-scale representation is crucial for downstream tasks. For example, Functa~\cite{dupont2022data} achieves CIFAR-10 classification accuracies of 68.3\%, 68.3\%, and 66.7\% for latent sizes of 256, 512, and 1024, respectively, indicating that relying solely on larger global latents does not capture task-specific features effectively. Moreover, our method achieves significantly better classification performance compared to local-based approaches~\cite{bauer2023spatial} and pixel-based methods~\cite{liu2022swin,dosovitskiy2020image,he2016deep,xie2017aggregated}, demonstrating that effectively leveraging both local and global information is crucial for downstream tasks. In addition to its superior accuracy, our approach is also highly efficient. For instance, our higher-resolution CelebA-HQ ($64^2$) model exhibits superior computational performance and scalability compared to the lower-resolution CIFAR-10 ($32^2$) dataset used in Spatial Functa. The high computational demands of Spatial Functa make it challenging to extend to 3D data, whereas our framework remains computationally feasible. These results reinforce the importance of balancing global context with local details to obtain more expressive and robust latent representations (see Experiments).  

We also explored alternative designs, such as using different activation functions for varying latent resolutions. This resulted in a modulated activation function expressed as:  
\[
\Sigma_{s=1}^{M} \lambda_s \sin{(W_0 (Wx + m_{s \times s}))},
\]  
where \( m_{s \times s} \) is the modulation latent with a size of \( s \times s \), combined with a scaling factor \( \lambda_s \) to emphasize local latents. However, this modification led to a marginal reduction in reconstruction performance compared to the hierarchical design. Furthermore, we tested removing the hierarchical modulation module and instead introduced a learnable scaling modulation factor in addition to the shift modulations. While this approach simplifies the architecture, it resulted in a noticeable drop in performance, underscoring the importance of hierarchical modulation in achieving robust and high-quality representations. 

Overall, our reconstruction model, LIFT, is distinguished by its speed and scalability across various resolutions, making it highly effective for both low and high-resolution data. Moreover, LIFT converges rapidly within just a few iterations, significantly shortening training durations. Notably, the ReLIFT variant accelerates convergence even further, proving especially beneficial when training time is constrained. Consequently, our framework is designed to tackle the challenges of large-scale datasets with LIFT, as well as address single-data scenarios using ReLIFT.
{
    \small
    \bibliographystyle{ieeenat_fullname}
    \bibliography{main}
}


\end{document}